\documentclass[runningheads]{llncs}

\usepackage{eccv}

\usepackage{graphicx}
\usepackage{booktabs}

\usepackage{tikz}
\usepackage{comment}
\usepackage{amsmath,amssymb} % define this before the line numbering.
\usepackage{color}

\usepackage[accsupp]{axessibility}  % Improves PDF readability for those with disabilities.

\usepackage{amsfonts}
\usepackage{subcaption}
\usepackage{bm}

\usepackage{verbatim}
\usepackage{caption}
\usepackage{wrapfig}
\usepackage{diagbox}
\usepackage{multirow}
\usepackage{color}
\definecolor{myred}{rgb}{1,0,0} 
\usepackage{cite}
\usepackage{enumitem}
\usepackage{makecell}

\usepackage{xcolor} 
\usepackage{url}
\usepackage{color, colortbl}

\makeatletter
\newcommand{\@chapapp}{\relax}%
\makeatother
\usepackage[title,header]{appendix}

\definecolor{cyberpink}{RGB}{235,110, 152}
\definecolor{lbrown}{RGB}{230,100, 20}
\definecolor{lime}{RGB}{205, 237, 250}
\definecolor{pink}{RGB}{252, 237, 246}
\definecolor{ourcolor}{RGB}{226,219,235}
\definecolor{purple}{RGB}{99,51,151}

\definecolor{lime}{RGB}{205, 237, 250}
\definecolor{pink}{RGB}{252, 237, 246}

\definecolor{rlime}{RGB}{205, 237, 250}
\definecolor{rlemon}{RGB}{255, 250, 186}

\definecolor{lemon}{RGB}{218, 243, 245}

\newcommand{\gauss}[0]{3DGS }
\newcommand{\dc}[0]{3DGS-No-SH }
\newcommand{\ours}[0]{CompGS }
\newcommand{\tandt}[0]{Tanks\&Temples }
\newcommand{\mipnerf}[0]{Mip-NeRF360 }
\newcommand{\kmeans}[0]{K-means }
\newcommand{\arkit}[0]{ARKit-200 }
\newcommand{\ignore}[1]{}

\usepackage[pagebackref,breaklinks,colorlinks,citecolor=eccvblue]{hyperref}

\usepackage{orcidlink}

\begin{document}

% ---------------------------------------------------------------
\title{CompGS: Smaller and Faster Gaussian Splatting with Vector Quantization} 

\titlerunning{CompGS: Smaller and Faster Gaussian Splatting}

\author{
K L Navaneet$^{*}\quad$Kossar Pourahmadi Meibodi$\thanks{Equal contribution}\quad$ \\Soroush Abbasi Koohpayegani $\quad$ Hamed Pirsiavash$\quad$   \\
}

\institute{University of California, Davis\\
\email{\{nkadur,kmeibodi,soroush,hpirsiav\}@ucdavis.edu}}

\authorrunning{Navaneet*, Pourahmadi Meibodi*, Abbasi Koohpayegani and Pirsiavash}

\maketitle

\begin{abstract}
3D Gaussian Splatting (3DGS) is a new method for modeling and rendering 3D radiance fields that achieves much faster learning and rendering time compared to SOTA NeRF methods. However, it comes with a drawback in the much larger storage demand compared to NeRF methods since it needs to store the parameters for several 3D Gaussians. We notice that many Gaussians may share similar parameters, so we introduce a simple vector quantization method based on K-means to quantize the Gaussian parameters while optimizing them. Then, we store the small codebook along with the index of the code for each Gaussian. We compress the indices further by sorting them and using a method similar to run-length encoding. Moreover, we use a simple regularizer to encourage zero opacity (invisible Gaussians) to reduce the storage and rendering time by a large factor through reducing the number of Gaussians. We do extensive experiments on standard benchmarks as well as an existing 3D dataset that is an order of magnitude larger than the standard benchmarks used in this field. 
We show that our simple yet effective method can reduce the storage cost for 3DGS by $40\times$ to $50\times$ and rendering time by $2\times$ to $3\times$ with a very small drop in the quality of rendered images. Our code is available here: 
{\href{https://github.com/UCDvision/compact3d}{https://github.com/UCDvision/compact3d}}

\end{abstract}

\section{Introduction}
\label{sec:intro}
\vspace{-.1in}

Recently, we have seen great progress in radiance field methods to reconstruct a 3D scene using multiple images captured from multiple viewpoints. NeRF~\cite{mildenhall2020nerf} is probably the most well-known method that employs an implicit neural representation to learn the radiance field using a deep model. Although very successful, NeRF methods are very slow to train and render. Several methods have been proposed to solve this problem; however, they usually come with some cost in the quality of the rendered images.

\begin{figure*}[t!]
    \centering
    \hspace{.2in}
    \begin{minipage}{.48\linewidth}
        \quad
        \begin{center}
          \includegraphics[width=\linewidth]{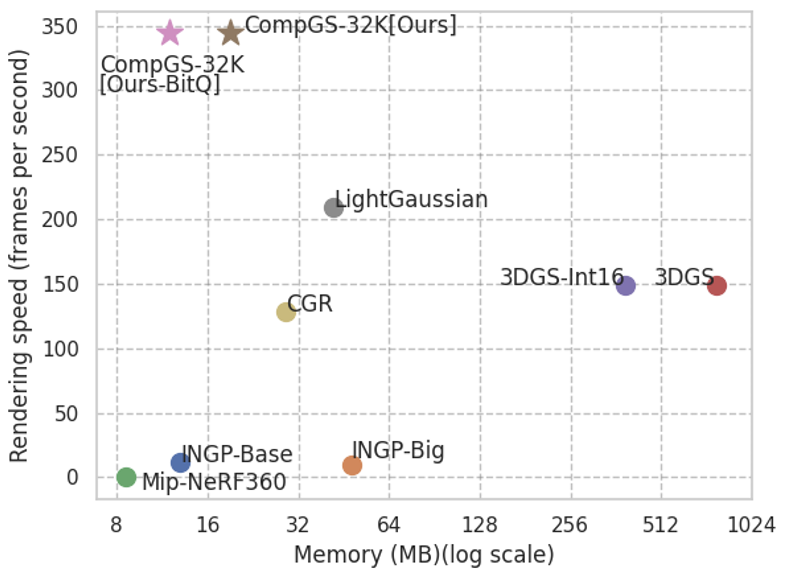}
        \end{center}
        % \vspace{-.2in}
           % \caption{\textbf{Results on ARKit dataset.} \dc fails to reconstruct well in several images while \ours is nearly identical to \gauss with a large reduction in model size.}
           
    \end{minipage}
    \hspace{-.2in}
    \begin{minipage}{.48\linewidth}
        \begin{center}
          \includegraphics[width=.6\linewidth]{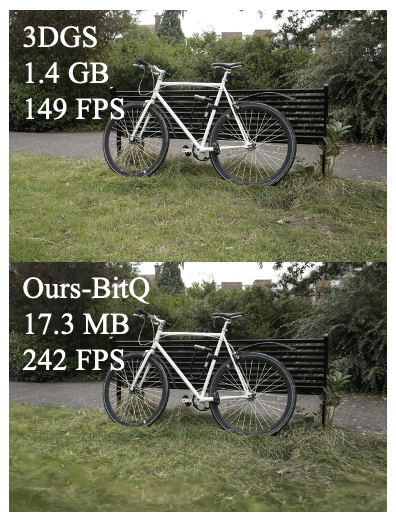}
        \end{center}
        % \vspace{-.2in}
           % \caption{\textbf{Results on ARKit dataset.} \dc fails to reconstruct well in several images while \ours is nearly identical to \gauss with a large reduction in model size.}
           
    \end{minipage}
    \caption{\textbf{Inference speed vs. memory comparison.} All methods except INGP achieve comparable PSNR that are reported in Table \ref{tab:comp_sota}. CompGS, our compressed version of 3DGS, maintains the speed and performance of \gauss while reducing its size to the levels of NeRF based approaches. We achieve around $45\times$ compression and $2.5\times$ inference speed up with little drop in performance (CompGS-32K). A bit quantized version of this (Ours-BitQ) compresses it further to a total compression of $65\times$ with hardly noticeable difference in quality.}
    \label{fig:all}
  \vspace{-.2in}
\end{figure*}

The Gaussian Splatting method (3DGS) ~\cite{kerbl20233d} is a new paradigm in learning radiance fields. The idea is to model the scene using a set of Gaussians. Each Gaussian has several parameters including its position in 3D space, covariance matrix, opacity, color, and spherical harmonics of the color that need to be learned from multiple-view images. 
Thanks to the simplicity of projecting 3D Gaussians to the 2D image space and rasterizing them, 3DGS is significantly faster to both train and render compared to NeRF methods. 
This results in real-time rendering of the scenes on a single GPU (ref. Fig.~\ref{fig:all}). Additionally, unlike the implicit representations in NeRF, the 3D structure of the scene is explicitly stored in the parameter space of the Gaussians. This enables many operations including editing the 3D scene directly in the parameter space.

One of the main drawbacks of the 3DGS method compared to NeRF variants is that 3DGS needs at least an order of magnitude more parameters compared to NeRF. This increases the storage and communication requirements of the model and its memory at the inference time, which can be very limiting in many real-world applications involving smaller devices. For instance, the large memory consumption may be prohibitive in storing, communicating, and rendering several radiance field models on AR/VR headsets.

We are interested in compacting 3DGS representations without sacrificing their rendering speed to enable their usage in various applications including low-storage or low-memory devices and AR/VR headsets. Our main intuition is that several Gaussians may share some of their parameters (e.g. covariance matrix). Hence, we simply vector-quantize parameters while learning and store the codebook along with the index for each Gaussian. This can result in a huge reduction in the storage. Also, it can reduce the memory footprint at the rendering time since the index can act as a pointer to the correct code freeing the memory needed to replicate those parameters for all Gaussians. 

\begin{figure*}
\begin{center}
\includegraphics[width=1\linewidth]{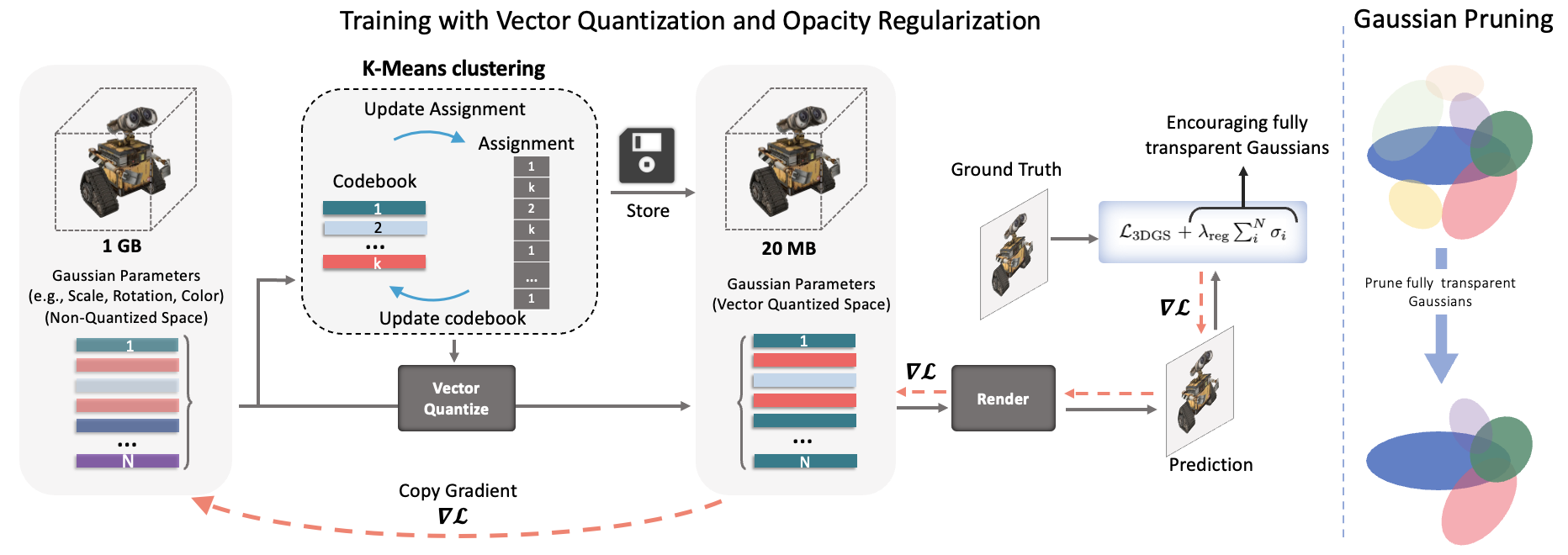}
\end{center}
 \vspace{-.2in}
   \caption{\textbf{Overview of \ours vector quantization:} We compress \gauss using vector quantization of the parameters of the Gaussians. The quantization is performed along with the training of the Gaussian parameters. Considering each Gaussian as a vector, we perform \kmeans clustering to represent the $N$ Gaussians in the model with $k$ cluster centers (codes). Each Gaussian is then replaced by its corresponding code for rendering and loss calculation. The gradients wrt centers are copied to all the elements in the corresponding cluster and the non-quantized versions of the parameters are updated. Only the codebook and code assignments for each Gaussian are stored and used for inference. 
   To further reduce the storage and inference time, we regularize opacity in the loss to encourage fully transparent Gaussians.
   \ours maintains the real-time rendering property of \gauss while compressing it by an order of magnitude.}
   \label{fig:teaser}
 \vspace{-.2in}
\end{figure*}

To this end, we use simple \kmeans algorithm to vector quantize the parameters at the learning time. Inspired by various quantization-aware learning methods in deep learning~\cite{rastegari2016xnornet}, we use the quantized model in the forward pass while updating the non-quantized model in the backward pass. To reduce the computation overhead of running K-means, we update the centroids in each iteration, but update the assignments less frequently (e.g., once every 100 iterations) since it is costly. Moreover, since the Gaussians are a set of non-ordered elements, we compress the representation further by sorting the Gaussians based on one of the quantized indices and storing them using the Run-Length-Encoding (RLE) method. 
Furthermore, we employ a simple regularizer to promote zero opacity (essentially invisible Gaussians), resulting in a significant reduction in storage and rendering time by reducing the number of Gaussians.
Our final model is $40\times$ to $50\times$ smaller and $2\times$ to $3\times$ faster during rendering compared to 3DGS.

\vspace{-.1in}
\section{Related Work}
\label{sec:related}
\vspace{-.1in}
{\bf Novel-view synthesis methods:} Early deep learning techniques for novel-view synthesis used CNNs to estimate blending weights or texture-space solutions~\cite{zhou2016view,flynn2016deepstereo,hedman2018deep,thies2019deferred,riegler2020free}. However, the use of CNNs faced challenges with MVS-based geometry and caused temporal flickering. Volumetric representations began with Soft3D ~\cite{penner2017soft}, and subsequent techniques used deep learning with volumetric ray-marching ~\cite{henzler2019escaping,sitzmann2019deepvoxels}. Mildenhall et al.~introduced Neural Radiance
Fields (NeRFs)~\cite{mildenhall2020nerf} to improve the quality of synthesized novel views, but the use of a large Multi-Layer Perceptron (MLP) as the backbone and dense sampling slowed down the process a lot. Successive methods sought to balance quality and speed, with Mip-NeRF360 achieving top image quality ~\cite{barron2022mipnerf360}. Recent advances prioritize faster training and rendering via spatial data structures, encodings, and MLP adjustments ~\cite{plenoxels, yu2021plenoctrees, hedman2021snerg, chen2023mobilenerf, garbin2021fastnerf, reiser2021kilonerf, wu2022snisr, takikawa2022variable, mueller2022instant}. Notable methods, like InstantNGP \cite{mueller2022instant}, use hash grids and occupancy grids for accelerated computation with a smaller MLP, while Plenoxels~\cite{plenoxels} entirely forgo neural networks, relying on Spherical Harmonics for directional effects. Despite impressive results, challenges in representing empty space, limitations in image quality, and rendering speed persist in NeRF methods. In contrast, 3DGS \cite{kerbl20233d} achieves superior quality and faster rendering without implicit learning ~\cite{barron2022mipnerf360}. However, the main drawback of 3DGS is its increased storage compared to NeRF methods which may limit its usage in many applications such as edge devices. We are able to keep the quality and fast rendering speed of 3DGS method while providing reduced model storage by applying a vector quantization scheme to Gaussian parameters.

\noindent\textbf{Bit quantization:} Reducing the number of bits to represent each parameter in a deep neural network is a commonly used method to quantize models~\cite{jacob2018quantization, han2015deep, krishnamoorthi2018quantizing} that result in smaller memory footprints. Representing weights in $64$ or $32$-bit formats may not be crucial for a given task, and a lower-precision quantization can lead to memory and speed improvements. Dettmers et al.~\cite{dettmers2022llm} show 8-bit quantization is sufficient for large language models. In the extreme case, weights of neural networks can be quantized using binary values. XNOR~\cite{rastegari2016xnor} examines this extreme case by quantization-aware training of a full-precision network that is robust to quantization transformations.

\noindent\textbf{Vector quantization:} Vector quantization (VQ) \cite{gray1984vector, gersho2012vector, equitz1989new, gong2014compressing} is a lossy compression technique that converts a large set of vectors into a smaller codebook and represents each vector by one of the codes in the codebook. As a result, one needs to store only the code assignments and the codebook instead of storing all vectors. 
VQ is used in many applications including image compression \cite{cosman1993using}, video and audio codec \cite{lee1995motion, makhoul1985vector}, compressing deep networks \cite{cho2023edkm,gong2014compressing}, and generative models \cite{van2017neural, gu2022vector, razavi2019generating}. We apply a similar method to compressing 3DGS models. 

\noindent\textbf{Deep model compression.} Model compression tries to reduce the storage size without changing the accuracy of original models. Model compression techniques can be divided to 1) model pruning \cite{gupta2015deep, han2015deep, vanhoucke2011improving, wen2016learning} that aims to remove redundant layers of neural networks; 2) weight quantization \cite{nooralinejad2023pranc, jacob2018quantization, krishnamoorthi2018quantizing}, and 3) knowledge distillation \cite{ba2013deep,hinton2015distilling,chen2017learning,polino2018model, abbasi2020compress}, in which a compact student model is trained to mimic the original teacher model. Some works have applied these techniques to volumetric radiance fields \cite{deng2023compressing, li2022streaming, yu2021plenoctrees}. For instance, TensoRF \cite{chen2022tensorf} decompose volumetric representations via low-rank approximation. 

\noindent\textbf{Compression for 3D scene representation methods.} Since NeRF relies on dense sampling of color values and opacity, the computational costs are significant. To increase efficiency, methods adopt different data structures such as trees \cite{wang2022fourier, yu2021plenoctrees}, point clouds \cite{xu2022point, peng2021shape}, and grids \cite{chen2022tensorf, plenoxels, mueller2022instant, schwarz2022voxgraf, sun2022direct, takikawa2022variable}. With grid structures training iterations can be completed in a matter of minutes. However, dense 3D grid structures may require substantial amounts of memory. Several methods have worked on reducing the size of such volumetric grids \cite{tang2022compressible, chen2022tensorf, takikawa2022variable, mueller2022instant}. Instant-NGP \cite{mueller2022instant} uses hash-based multi-resolution grids. VQAD \cite{takikawa2022variable} replaces the hash function with codebooks and vector quantization. 
Another line of work decomposes 3D grids into lower dimensional components, such as planes and vectors, to reduce the memory requirements~\cite{chen2022tensorf, tang2022compressible, huang2022pref}. Despite reducing the time and space complexity of the 3D scenes, their sizes are still larger than MLP-based methods. VQRF \cite{li2023compressing} compresses volumetric grid-based radiance fields by adopting the VQ strategy to encode color features into a compact codebook. \newline \indent While we also employ vector quantization, we differ from the above approaches in the method employed for novel view synthesis. Unlike the NeRF based approaches described above, we aim to compress \gauss which uses a collection of 3D Gaussians to represent the 3D scene and does not contain grid like structures or neural networks. We also achieve a significant amount of compression by regularizing and pruning the Gaussians based on their opacity. 

\noindent {\bf Concurrent works:} Some very recent works developed concurrently to ours \cite{niedermayr2023compressed,fan2023lightgaussian,lee2023compact,morgenstern2023compact,girish2023eagles} also propose vector quantization and pruning based methods to compress 3D Gaussian splat models. LightGaussian~\cite{fan2023lightgaussian} uses importance based Gaussian pruning and distillation and vector quantization of spherical harmonics parameters. Similarly, CGR~\cite{lee2023compact} masks Gaussians based on their volume and transparency to reduce the number of Gaussians and uses residual vector quantization for scale and rotation parameters. In CGS~\cite{niedermayr2023compressed}, highly sensitive parameters are left non-quantized while the less sensitive ones are vector quantized. 

% \vspace{-.1in}
\section{Method}
\label{sec:method}
% \vspace{-.1in}
Here, we briefly describe the 3DGS~\cite{kerbl20233d} method for learning and rendering 3D scenes and explain our vector quantization approach for compressing it. \\

\noindent \textbf{Overview of 3DGS:} 3DGS models a scene using a collection of 3D Gaussians. A 3D
Gaussian is parameterized by its position and covariance matrices in the 3D space. 
$G(x) = e^{-\frac{1}{2}(x-\mu)^T \Sigma^{-1} (x-\mu)}$
where $x-\mu$ is the position vector, $\mu$ is the position, and $\Sigma$ is the 3D covariance matrix of the Gaussian. Since the covariance matrix needs to be positive definite, it is factored into its scale ($S$) and rotation ($R$) matrices as $\Sigma = RSS^TR^T$ for easier optimization.
In addition, each  Gaussian has an opacity parameter $\sigma$. Since the color of the Gaussians may depend on the viewing angle, the color of each Gaussian is modeled by a Spherical Harmonics (SH) of order 3 in addition to a DC component.

Given a view-point, the collection of 3D Gaussians is efficiently rendered in a differentiable manner to get a 2D image by $\alpha$-blending of anisotropic splats, sorting, and using a tile-based rasterizer. Color of a pixel is given by 
$C = \sum_i c_i \alpha_i \prod_{j=1}^{i-1}(1-\alpha_j)$ 
where $c_i$ is the color of the $i^{th}$ Gaussian and $\alpha_i$ is
the product of the value of the Gaussian at that point and its learned opacity $\sigma_i$. At the training time, 3DGS minimizes the loss between the groundtruth and rendered images in the pixel space. 
The loss is $\ell_1$ loss plus an SSIM loss in the pixel space. 3DGS initializes the optimization by a point cloud achieved by a standard SfM method and iteratively prunes the ones with small opacity and adds new ones when the gradient is large. 3DGS paper shows that it is extremely fast to train and is capable of real-time rendering while matching or outperforming SOTA NeRF methods in terms of rendered image quality. \\

\noindent{\bf Compression of 3DGS:}

We compress the parameters of 3DGS using vector quantization aware training and reduce the number of Gaussians by regularizing the opacity parameter. 

\noindent \textbf{Vector quantization: }
3DGS requires a few million Gaussians to model a typical scene. With $59$ parameters per Gaussian, the storage size of the trained model is an order of magnitude larger than most NeRF approaches (e.g., Mip-NeRF360~\cite{barron2022mipnerf360}). This makes it inefficient for some applications including edge devices. We are interested in reducing the number of parameters. Our main intuition is that many Gaussians may have similar parameter values (e.g., covariance). Hence, we use simple vector quantization using \kmeans algorithm to compress the parameters. Fig.~\ref{fig:teaser} provides an overview of our approach.

Consider a 3DGS model with $N$ Gaussians, each with a $d$ dimensional parameter vector.
We run \kmeans algorithm to cluster the vectors into $K$ clusters. Then, one can store the model using $K$ vectors of size $d$ and $N$ integer indices (one for each Gaussian). Since $N >> K$, this method can result in a large compression ratios. In a  typical scene, $N$ is a few millions while $K$ is a few thousands.

However, clustering the model parameters after training results in performance degradation, hence, we perform quantization aware training to ensure that the parameters are amenable to quantization. In learning 3DGS, we store the non-quantized parameters. In the forward pass of learning 3DGS, we quantize the parameters and replace them with the quantized version (centroids) to do the rendering and calculate the loss. Then, we do the backward pass to get the gradients for the quantized parameters and copy the gradients to the non-quantized parameters to update them. We use straight-through estimator proposed in STE~\cite{bengio2013estimating}. After learning, we discard the non-quantized parameters and keep only the codebook and indices of the codes for Gaussians.

Since the number of Gaussians $N$ is typically in millions, cost of performing \kmeans at every iteration of training can be prohibitively high. 
\kmeans has two steps: updating centroids given assignments, and updating assignments given centroids. We note that the latter is more expensive while the former is a simple averaging. Hence, we update the centroids after each iteration and update the assignments once every $t$ iterations. We observe that
the modified approach works well even for values of $t$ as high as $500$. This is crucial in limiting the training time of the method.

Performing a single \kmeans for the whole $d$ dimensional parameters requires a huge codebook since the different parameters of the Gaussian are not necessarily correlated. Hence, we group similar types of parameters, e.g., all rotation matrices, together and cluster them independently to learn a separate codebook for each. This requires storing multiple indices for each Gaussian. 
In our main method, we quantize DC component of color, spherical harmonics, scale, and rotation parameters separately, resulting in $4$ codebooks. We do not quantize opacity parameter since it is a single scalar and do not quantize the position of the Gaussians since sharing them results in overlapping Gaussians.

Since the indices are integer values, we use fewer number of bits compared to the original parameters to store each. Moreover, 3DGS models the scene as a set of order-less Gaussians. Hence, we sort the Gaussians based on one of the indices, e.g., rotation, so that Gaussians using the same code appear together in the list. Then, for that index, instead of storing $n$ integers, we store how many times each code appears in the list, reducong the storage from $n$ integers to $k$ integers. This is similar to run-length-encoding for data compression. \\

\noindent \textbf{Opacity Regularization: }
Some parameters like position of the Gaussians cannot be quantized easily, so as shown in Table~\ref{tab:kmeans_breakdown} after quantization, they dominate the memory(more than 80\% of memory). This means quantization cannot improve the compression any further. One way to compress 3DGS more is to reduce the number of Gaussians. Interestingly, this reduction comes with a bi-product that is increase in inference speed.
We know that very small values of opacity ($\sigma$) correspond to transparent or nearly invisible Gaussians. Hence, inspired by training sparse models, we add $\ell_1$ norm of the opacity to the loss as a regularizer to encourage zero values for opacity. 
Therefore, the final loss becomes: $\mathcal{L} = \mathcal{L}_{\textrm{3DGS}} + \lambda_{\textrm{reg}}\sum_i^N \sigma_i$, where $\mathcal{L}_{\textrm{3DGS}}$ is the original loss of 3DGS with or without quantization and $\lambda_{\textrm{reg}}$ controls the sparsity of opacity. Finally, similar to the original 3DGS, we remove the Guassians with opacity smaller than a threshold, resulting in significant reduction in storage and inference time.

% \vspace{-.1in}
\section{Experiments}
\label{sec:exp}
% \vspace{-.1in}
\noindent \textbf{Implementation details:} For all our experiments, we use the publicly available 
official code repository~\cite{gsplatcode} of 3DGS ~\cite{kerbl20233d} provided by its authors.
There are no changes in the hyperparameters used for training compared to 3DGS. The Gaussian
parameters are trained without any vector quantization till $20K$ iterations and \kmeans quantization is
used for the remaining $10K$ iterations. A standard \kmeans iteration involves 
distance calculation between all elements (Gaussian parameters) and all cluster centers followed by assignment
to the closest center. The centers are then updated using new cluster assignments and the loop is repeated.
We use just $1$ such \kmeans iteration in our experiments once every $100$ iterations
till iteration $25K$ and keep the assignments constant thereafter till the last iteration, $30K$.
The \kmeans cluster centers are 
updated using the non-quantized Gaussian parameters after each iteration of training.  
The covariance (scale and rotation) and color
(DC and harmonics) components of each Gaussian is vector quantized while position (mean) and opacity parameters are not quantized. Additional results with different parameters being quantized
are provided in Table \ref{tab:kmeans_variants}. Unless mentioned differently, we use a codebook of size $4096$ for the color 
and $16384$ (\ours $16K$) for the covariance parameters. The scale parameters of covariance are quantized before applying 
the exponential activation on them. Similarly, quaternion based rotation parameters are quantized before 
normalization. For opacity regularization, we use $\lambda_{\textrm{reg}}=10^{-7}$ from iterations $15K$ to $20K$
along with opacity based pruning every $1000$ iterations and remove regularization thereafter. 
All experiments were run on a single RTX-$6000$ GPU. \\

\begin{table*}[t]
    %\vspace{-0.5cm}
    \caption{\textbf{Comparison with SOTA methods for novel view synthesis.} 
    3DGS~\cite{kerbl20233d} performs comparably or outperforms the best of the NeRF based approaches while maintaining a high rendering speed during inference. Trained NeRF models are significantly smaller than \gauss since NeRFs are parameterized using neural networks while \gauss requires storage of parameters of millions of 3D Gaussians. \ours is a vector quantized version of \gauss that maintains the speed and performance advantages of \gauss while being \bm{$40\times$} to \bm{$50\times$} \textbf{smaller}. 
    \ours 32K BitQ is the post-training bit quantized version of \ours 32K, in which position parameters are 16-bits, opacity is 8 bits, and the rest are 32 bits.
    $^*$Reproduced using official code. $^\dag$ Reported from 3DGS~\cite{kerbl20233d}. Our timings for \gauss and \ours are reported using a RTX6000 GPU while those with $^\dag$ used A6000 GPU. We boldface entries for emphasis. Please see the Appendix for results on Deep Blending dataset.
    }
    \centering
    \scalebox{0.78}{
    \begin{tabular}{lcccccc|cccccc}
    \toprule
     & \multicolumn{6}{c}{Mip-NeRF360} & \multicolumn{6}{c}{Tanks\&Temples}  \\ 
    Method & SSIM$^\uparrow$ & PSNR$^\uparrow$ & LPIPS$^\downarrow$ & FPS & \makecell{Mem \\ (MB)} & \makecell{Train \\ Time(m)} & SSIM$^\uparrow$ & PSNR$^\uparrow$ & LPIPS$^\downarrow$ & FPS & \makecell{Mem \\ (MB)} & \makecell{Train \\ Time(m)} \\
    \midrule
    Plenoxels$^\dag$~\cite{plenoxels} & 0.626 & 23.08 & 0.463 & 6.79 & 2,100 & 25.5 & 0.719 & 21.08 & 0.379 & 13.0 & 2300 & 25.5   \\ % & 10.3 &  2.4 GB\\
    INGP-Base$^\dag$~\cite{mueller2022instant} & 0.671 & 25.30 & 0.371 & 11.7 & 13 & 5.37 & 0.723 & 21.72 & 0.330 & 17.1 & 13 & 5.26  \\ % & 10.7 &  13 MB\\
    INGP-Big$^\dag$~\cite{mueller2022instant} & 0.699 & 25.59 & 0.331 & 9.43 & 48 & 7.30 & 0.745 & 21.92 & 0.305 & 14.4 & 48 & 6.59  \\ % & 8.86 &  48 MB\\
    M-NeRF360$^\dag$~\cite{barron2022mipnerf360} & 0.792 & 27.69 & 0.237 & 0.06 & 8.6 & 48h & 0.759 & 22.22 & 0.257 & 0.14 & 8.6 & 48h  \\ % & 0.09 &  8.6 MB\\
    \gauss$^\dag$~\cite{kerbl20233d} & 0.815 & 27.21 & 0.214 & 134 & 734 & 41.3 & 0.841 & 23.14 & 0.183 & 154 & 411 & 26.5   \\ % & 142  &  607 MB\\
    \gauss$^*$~\cite{kerbl20233d}   & 0.813 & 27.42 & 0.217 & $\bm{149}$ & $\bm{778}$ & 21.6 & 0.844 & 23.68 & 0.178 & $\bm{206}$ & $\bm{433}$ & 12.2  \\ 
    LigthGaussian~\cite{fan2023lightgaussian}    & 0.805 & 27.28 & 0.243 & 209 & 42    & - & 0.817 & 23.11 & 0.231 & 209 & 22    & -  \\
    CGR~\cite{lee2024compact}              & 0.797 & 27.03 & 0.247 & 128 & 29.1  & - & 0.831 & 23.32 & 0.202 & 185 & 20.9  & -  \\
    CGS~\cite{niedermayr2024compressed}    & 0.801 & 26.98 & 0.238 & -   & 28.8 & - & 0.832 & 23.32 & 0.194 & -   & 17.28 & -  \\
    \rowcolor{ourcolor}
    \ours 16K        & 0.804 & 27.03 & 0.243 & $\bm{346}$ & $\bm{18}$ & 22.8 & 0.836 & 23.39 & 0.200 & $\bm{479}$ & $\bm{12}$  & 15.6  \\
    \rowcolor{ourcolor}
    \ours 32K        & 0.806 & 27.12 & 0.240 & 344 & 19 & 29.4 & 0.838 & 23.44 & 0.198 & 475 & 13  & 20.6 \\
    \rowcolor{ourcolor}
    \ours 32K BitQ   & 0.797 & 26.97 & 0.245 & $\bm{344}$ & $\bm{12}$ & 29.4 & 0.832 & 23.35 & 0.202 & $\bm{475}$ & $\bm{8}$  & 20.6 \\
    
    \bottomrule
    \end{tabular}
    }
    % \vspace{-0.5cm}
    \label{tab:comp_sota}
\end{table*}

\noindent \textbf{Datasets:} We primarily show results on three challenging real world datasets - 
Tanks\&Temples\cite{knapitsch2017tanks}, Deep Blending~\cite{hedman2018deep} and \mipnerf~\cite{barron2022mipnerf360}
containing two, two and nine scenes respectively.
Also, we provide results on a subset of the recently released DL3DV-10K dataset \cite{ling2023dl3dv} which contains 140 scenes. DL3DV-10K~\cite{ling2023dl3dv} is an annotated dataset with $10,510$ real-world scene-level videos. Out of these, $140$ scenes have been used to create a novel-view synthesis (NVS) benchmark, making it an order of magnitude larger than the typical NVS benchmarks. We use this NVS benchmark in our experiments. Additionally, we provide results on a subset of the large scale ARKit \cite{baruch2021arkitscenes}
dataset, called ARKit-200, which contains $200$ scenes. Details of this dataset is presented in the Appendix. \\ 

\noindent \textbf{Baselines:} As we propose a method (termed CompGS) 
for compacting 3DGS, we focus our comparisons with
3DGS and different baseline methods for compressing it. We consider bit quantization (denoted as Int-16/8/4 in 
results) and \gauss without the harmonic components for color (denoted as 3DGS-No-SH) as alternative parameter compression
methods. Bit-quantization is performed using the standard Absmax quantization~\cite{dettmers2022llm} technique.
Similarly, we consider several alternative approaches to reduce the number of Gaussians. 
Densification process in 3DGS increases the Gaussian count and is controlled by the gradient threshold (termed grad thresh) parameter and the 
frequency (freq) and iterations (iters) until densification is performed. The opacity threshold (min opacity) controls the pruning of transparent 
Gaussians. We modify these parameters in 3DGS to compress the model with as little drop in performance as possible. 
Additionally, Table ~\ref{tab:comp_sota} shows comparison with state-of-the-art NeRF 
approaches~\cite{barron2022mipnerf360,plenoxels,mueller2022instant}.
Mip-NeRF360~\cite{barron2022mipnerf360} achieves high performance comparable to \gauss while 
Plenoxels~\cite{plenoxels} and InstantNGP\cite{mueller2022instant} have high frame-rate for 
rendering and very low training time. InstantNGP and \mipnerf are also comparable in model size to our 
compressed model. \\

\noindent \textbf{Evaluation:} For a fair comparison, we use the same train-test split as 
\mipnerf\cite{barron2022mipnerf360} and 3DGS~\cite{kerbl20233d} and directly report the metrics for other
methods from 3DGS~\cite{kerbl20233d}. We also report our reproduced metrics for \gauss since we observe slightly better results compared to the ones in~\cite{kerbl20233d}. We report the standard evaluation 
metrics of SSIM, PSNR and LPIPS along with memory or compression ratio, rendering FPS and training time. The common practice is to report the average of PSNR across a set of images and scenes. 
However, this metric may be dominated by very accurate reconstructions (smaller errors) since it is based on the geometric average of the errors due to the log operation in PSNR calculation. Hence, for the larger ARKit dataset, we also report PSNR-AM for which we average the error across all images and scenes before calculating the PSNR. In comparing model sizes, we normalize all methods by dividing them by the size of our method to obtain compression ratio. \\ 

\begin{table*}[t]
    % \vspace{-0.2in}

    \caption{\textbf{Comparison of parameter compression methods for 3DGS.} We evaluate different baseline approaches for compressing the parameters of 3DGS without any reduction in the number of Gaussians. All memory values are reported as a ratio of the method with our smallest model. Our K-Means based vector quantization performs favorably compared to all methods both in terms of novel view synthesis performance and compression. 
    Not quantizing the position values (Int-x no-pos) is crucial in bit quantization.
    Since harmonics constitute $76\%$ of each Gaussian, 3DGS-no-SH achieves a high level of compression. But \ours with only quantized harmonics achieves similar compression with nearly no loss in performance compared to \gauss.}
    \centering
    \scalebox{0.93}{
    \begin{tabular}{lccc|ccc|ccc|c}
    \toprule
          & \multicolumn{3}{c}{\mipnerf} & \multicolumn{3}{c}{\tandt} & \multicolumn{3}{c}{Deep Blending} \\
        Method & SSIM & PSNR & LPIPS & SSIM & PSNR & LPIPS & SSIM & PSNR & LPIPS & Mem\\
        \midrule
        \gauss                  & 0.813 & 27.42 & 0.217 & 0.844 & 23.68 & 0.178 & 0.899 & 29.49 & 0.246 & 20.0  \\  % & 21.2 \\
        \dc                     & 0.802 & 26.80 & 0.229 & 0.833 & 23.16 & 0.190 & 0.900 & 29.50 & 0.247 & 4.8  \\  % & 5  \\
        Post-train \kmeans 4K   & 0.768 & 25.46 & 0.266 & 0.803 & 22.12 & 0.226 & 0.887 & 28.61 & 0.268 & 1.7  \\  % &  \\
        \rowcolor{ourcolor}      
        \kmeans 4K Only-SH      & 0.811 & 27.25 & 0.223 & 0.842 & 23.57 & 0.183 & 0.902 & 29.60 & 0.246 & 4.8  \\  % & 1.8\\
        \rowcolor{ourcolor}      
        \kmeans 4K              & 0.804 & 26.97 & 0.234 & 0.836 & 23.31 & 0.194 & 0.904 & 29.76 & 0.248 & 1.7  \\  % & 1.8\\
        \rowcolor{ourcolor}      
        \kmeans 32K             & 0.808 & 27.16 & 0.228 & 0.840 & 23.47 & 0.188 & 0.903 & 29.75 & 0.247 & 1.8  \\  % & 1.9 \\
        \midrule      
        Int16                   & 0.804 & 27.25 & 0.223 & 0.836 & 23.56 & 0.185 & 0.900 & 29.49 & 0.247 & 10.0  \\  % & 10.6 \\ % fp16 for all
        Int8 no-pos             & 0.812 & 27.38 & 0.219 & 0.843 & 23.67 & 0.180 & 0.900 & 29.47 & 0.247 & 5.8  \\  % & 6.1  \\ % all except xyz  
        Int8                    & 0.357 & 14.41 & 0.629 & 0.386 & 12.37 & 0.625 & 0.709 & 21.58 & 0.457 & 5.0  \\  % & 5.3 \\
        Int4 no-pos             & 0.489 & 17.42 & 0.525 & 0.488 & 12.94 & 0.575 & 0.746 & 19.90 & 0.446 & 3.4  \\  % & 3.6  \\  % all except xyz
        \dc Int16               & 0.789 & 26.59 & 0.237 & 0.826 & 23.04 & 0.198 & 0.900 & 29.50 & 0.248 & 2.4  \\  % & 2.5 \\
        \rowcolor{ourcolor}
        \kmeans 4K, Int16       & 0.796 & 26.83 & 0.239 & 0.830 & 23.21 & 0.199 & 0.904 & 29.76 & 0.248 & 1.0\\
        \bottomrule
    \end{tabular}
    }
    \vspace{-0.1in}
    \label{tab:comp_gauss}
\end{table*}

\noindent \textbf{Results:}
\label{subsec:results}
Comparison of our results with SOTA novel view synthesis approaches is shown in Table~\ref{tab:comp_sota}.
Our vector quantized method has a comparable performance to the non-quantized 3DGS with a small drop on MipNerf-360 and 
TandT datasets and a small improvement on the DB dataset. We additionally report results with post-training bit quantization of our model (\ours BitQ) where the position and opacity parameters are quantized to $16$ bits and $8$ bits respectively. 
The model memory footprint drastically reduces for \ours compared to \gauss, making it comparable to NeRF approaches. Our 
models are \bm{$65\times$} and \bm{$54\times$} \textbf{smaller} than 3DGS models on MipNerf-360 and TandT datasets respectively.
This reduces a big disadvantage of \gauss models and
makes them more practical. The compression achieved by \ours is impressive considering that more than
two-thirds of its memory is due to the non-quantized position and opacity parameters (refer table~\ref{tab:kmeans_breakdown}).
Additionally \ours maintains the other advantages of \gauss such as low inference memory usage and training time. 
while also \textbf{increasing its already impressive rendering FPS by} \bm{$2\times$} \textbf{to} \bm{$3\times$}.
A limitation of \ours compared to \gauss is the overhead in compute and training 
time introduced by the \kmeans clustering algorithm. This is compensated in part by the reduced compute and time due 
to the decrease in Gaussian count. \ours 16K variant requires marginally more time than \gauss while \ours 32K needs
$1.4\times$ to $1.7\times$ more training time. 
However, this is still orders of magnitude smaller than the high-quality NeRF based approaches like MipNerf-360. 
Per-scene evaluation metrics are in Appendix. Note that there are large differences in reproduced results for 3DGS across various works in the literature. We observe a median standard deviation of $0.05$dB for PSNR when the experiment is repeated $20$ times with several scenes having differences more than $0.4$dB across runs (refer Appendix). One must be careful when analyzing as these variations are often comparable to differences in performance between methods. \\

\begin{figure*}[t]
\vspace{-.1in}
\begin{center}
  \includegraphics[width=0.8\linewidth]{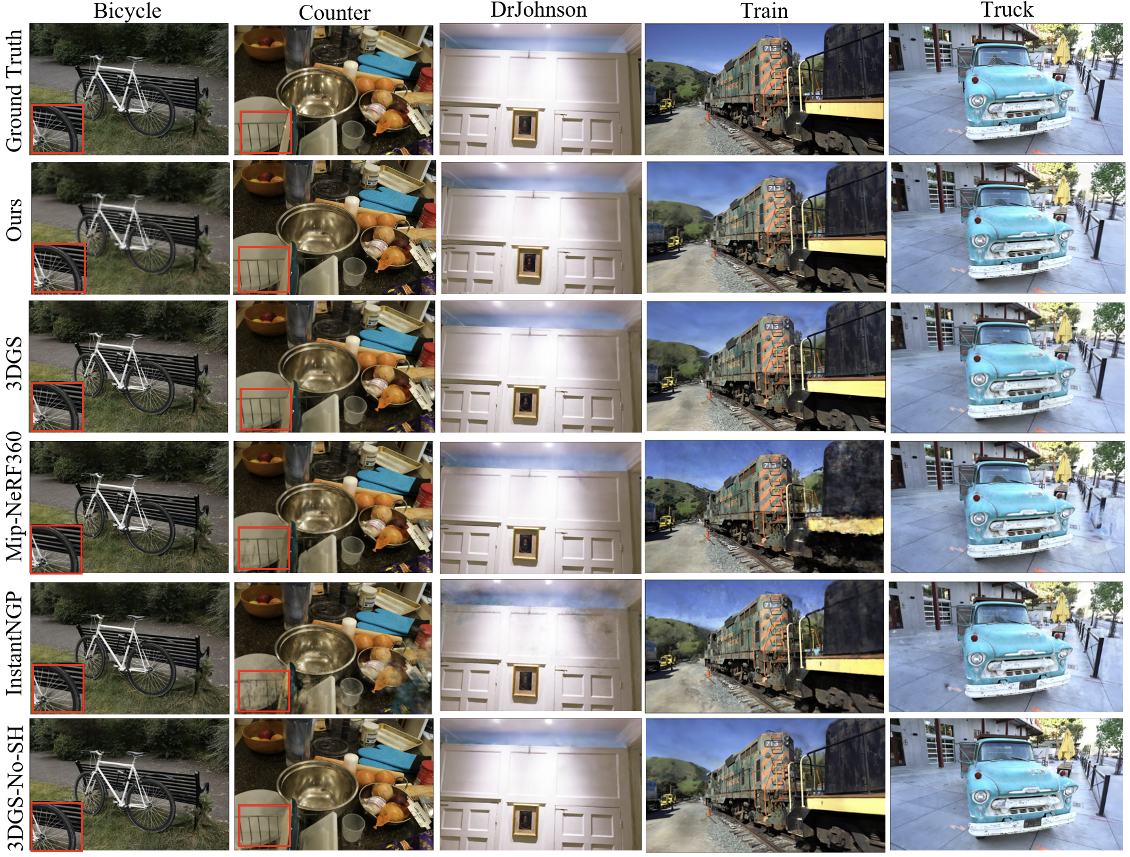}
\end{center}
   \vspace{-.15in}
   \caption{\textbf{Qualitative comparison of novel view synthesis approaches}. We visualize images from different scenes across datasets for SOTA NeRF, 3DGS, our \ours and the No-SH variant of \gauss. All methods based on \gauss have better reconstruction of finer details like spokes of the bicycle wheel compared to NeRF approaches. Both compressed versions \ours and \dc are similar in appearance to \gauss with no additional visually apparent errors.}
   \label{fig:vis_ours_vs_baselines}
 \vspace{-.2in}
\end{figure*}

We decouple our compression method into parameter and Gaussian count compression components and perform ablations on each of them.

\noindent \textbf{Comparison of parameter compression methods:} In Table~\ref{tab:comp_gauss}, we compare the proposed vector quantization
based compression against other baseline approaches for compressing 3DGS. Since the spherical harmonic 
components used for modeling color make up nearly three-fourths of all the parameters of each Gaussian,
a trivial compression baseline is to use a variant of \gauss with only the DC component for color and no
harmonics. This baseline (\dc) achieves a high compression with just $23.7\%$ of the original model size
but has a drop in performance. Our K-Means approach outperforms \dc while using less than half its 
memory. We also consider a variant of \ours with a single codebook for both SH and DC parameters (termed SH+DC)
with a larger codebook of size of $4096$. This has a marginal decrease in both memory and performance compared to 
default \ours suggesting that correlated parameters can be combined to reduce the number of indices to 
be stored. 

Fig.~\ref{fig:vis_ours_vs_baselines} shows qualitative comparison of \ours across multiple datasets with both SOTA approaches
and compression methods for \gauss. Both \ours and \dc are visually similar to 3DGS, preserving 
finer details such as the spokes of the bike and bars of dish-rack. Among NeRF approaches, \mipnerf
is closest in terms of quality to \gauss while InstantNGP trades-off quality for inference and training
speed. 

\begin{table*}[t]
    \caption{\textbf{Reducing number of Gaussians in 3DGS.} We evaluate different baseline approaches for compressing 3DGS by reducing the number of Gaussians. Gaussian count is proportional to model size. \ours performs favorably compared to all methods both in terms of novel view synthesis performance and compression. %Entries boldfaced for emphasis.
    % \vspace{-.15}
    }
    \centering
    \scalebox{0.8}{
    \begin{tabular}{lcccc|cccc|cccc}
    \toprule
               & \multicolumn{4}{c}{\mipnerf} & \multicolumn{4}{c}{\tandt} & \multicolumn{4}{c}{Deep Blending} \\
        Method & SSIM & PSNR & LPIPS & \#Gauss & SSIM & PSNR & LPIPS & \#Gauss & SSIM & PSNR & LPIPS & \#Gauss  \\ % & Mem\\
        \midrule
        \gauss                & 0.813 & 27.42 & 0.217 & $\bm{3.30M}$  & 0.844 & 23.68 & 0.178 & $\bm{1.83M}$ & 0.899 & 29.49 & 0.246 & $\bm{2.80M}$ \\  % & 21.2 \\
        Min Opacity           & 0.802 & 27.12 & 0.244 & 1.46M & 0.833 & 23.44 & 0.204 & 780K & 0.902 & 29.50 & 0.255 & 1.01M \\
        Densify Freq          & 0.794 & 26.98 & 0.255 & 1.07M & 0.832 & 23.36 & 0.206 & 709K & 0.902 & 29.76 & 0.258 & 844K \\
        Densify Iters         & 0.780 & 27.02 & 0.267 & 1.12M & 0.835 & 23.55 & 0.194 & 810K & 0.896 & 29.42 & 0.264 & 795K\\
        Grad Thresh           & 0.769 & 26.57 & 0.292 & 809K & 0.825 & 23.31 & 0.217  & 578K & 0.900 & 29.49 & 0.260 & 1.01M \\
        \rowcolor{ourcolor}
        Opacity Reg             & 0.813 & 27.42 & 0.227 & $\bm{845K}$ & 0.844 & 23.71 & 0.188 & $\bm{520K}$ & 0.905 & 29.73 & 0.249 & $\bm{554K}$ \\
        \bottomrule
    \end{tabular}
    }
    % \vspace{-0.3in}
    \label{tab:reg_ablation}
\end{table*}

\noindent All the above approaches are trained using $32$-bit precision for all Gaussian parameters. Post-training bit 
quantization of \gauss to $16$-bits reduces the memory by half with very little drop in performance. 
However, reducing the precision to $8$-bits results in a huge degradation of the model. 
This drop is due to the quantization of the position parameters of the Gaussians. Excluding them from 
quantization (denoted as Int8 no-pos) results in a model comparable to the $32$-bit variant. However, further
reduction to $4$-bits degrades the model even when the position parameters are not quantized. Note that 
bit quantization approaches offer significantly lower compression compared to \ours and they are a subset
of the possible solutions for our vector quantization method. Similar to 3DGS, \ours
has a small drop in performance when $16$-bit quantization is used. \\

\begin{table}[t]
%\vspace{-0.2in}
    \caption{\textbf{Comparison on ARKit-200 dataset.} It contains 200 scenes from the ARKit~\cite{arkitscenes} indoor scene understanding dataset (see the Appendix for details.). We report results for just the vector quantized version of CompGS.
    (left) \ours achieves a high level of compression with nearly identical metrics for view synthesis. (right) \dc fails to reconstruct well in several images while \ours is nearly identical to \gauss with a large reduction in model size.}
    \vspace{-0.1in}
    \begin{minipage}{.50\linewidth}
    \centering
    \scalebox{0.8}{
    \begin{tabular}{lccccc}
    \toprule
        Method & SSIM & PSNR & PSNR-AM & LPIPS &  Mem \\
        \midrule
        \gauss  & 0.909 & 25.76 & 20.73 & 0.226 & 20.0\\
        \dc     & 0.905 & 25.31 & 20.11 & 0.234 & 4.8 \\
        \ours   & 0.909 & 25.70 & 20.73 & 0.229 & 1.7\\
        \bottomrule
    \end{tabular}
    }
    \label{tab:kmeans_arkit}
    \end{minipage}%
    \begin{minipage}{.50\linewidth}
\begin{center}
  \includegraphics[width=\linewidth]{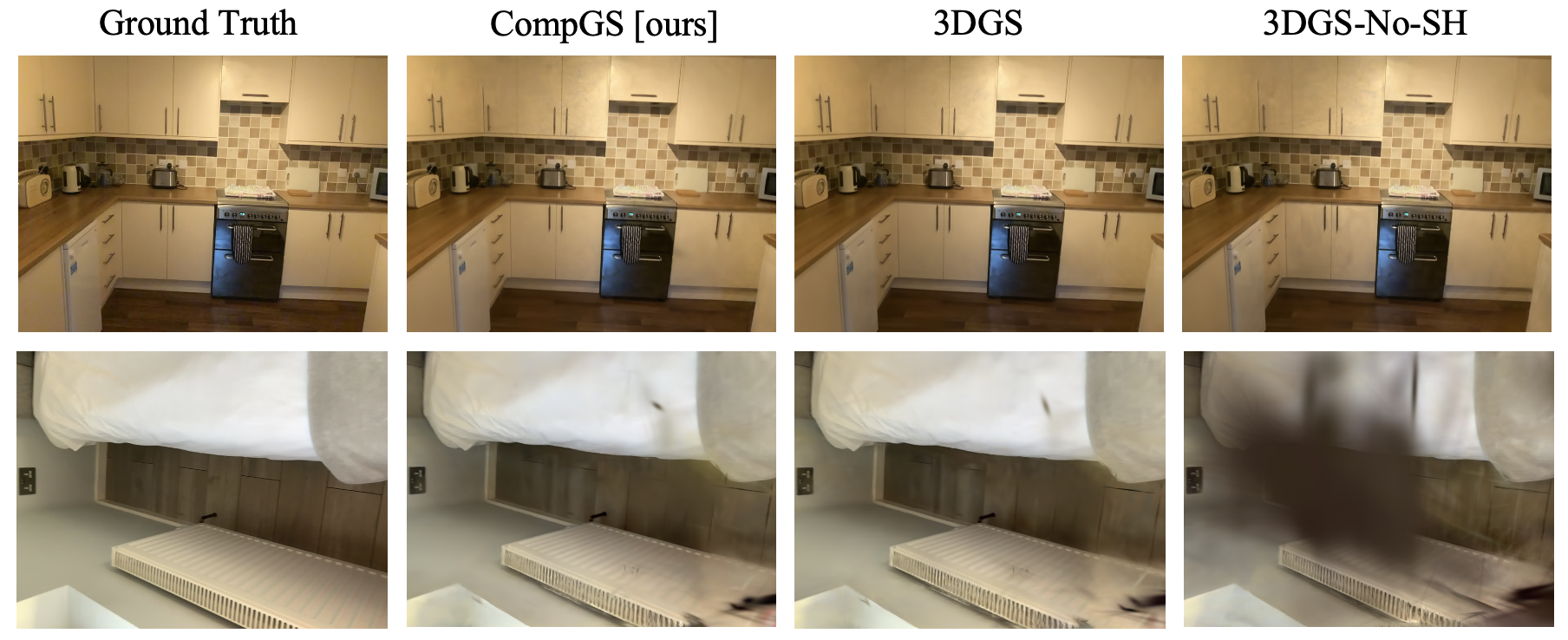}
\end{center}
\vspace{-.2in}
   \label{fig:arkit}
    \end{minipage}
% \vspace{-0.1in}
\end{table}

%\vspace{-.1in}
\begin{table}[]
    \centering
    \scalebox{0.9}{
    \begin{tabular}{lcccccc}
    \toprule
               & \multicolumn{5}{c}{DL3DV-140} \\
        Method & SSIM$^\uparrow$ & PSNR$^\uparrow$ & PSNR-AM$^\uparrow$ & LPIPS$^\downarrow$ & FPS & Mem(MB)  \\ % & Training Time (m)\\
        \midrule
        \gauss*  & 0.905 & 29.06 & 27.37 & 0.134  & 282 & 291      \\ % & 10.6 \\
        \ours 32K & 0.895 & 28.42 & 26.97 & 0.149 & 566 & 10          \\ % & 24.2 \\
        \bottomrule
    \end{tabular}
    }
    \caption{Results on the 140 scenes NVS benchmark of DL3DV-10K\cite{ling2023dl3dv} dataset. Similar to the results on the smaller benchmarks, \ours compresses 3DGS by nearly $30$ times with a small drop in reconstruction quality. * is our reproduced results. \vspace{-.25in}}
    \label{tab:kmeans_dl3dv}
\end{table}

\noindent \textbf{Comparison of Gaussian count compression methods:} In Table~\ref{tab:reg_ablation}, we compare the proposed opacity regularization
method for reducing the Gaussian count with baselines. In these baselines, we modify the 3DGS parameters to decrease densification and increase pruning 
and thus reduce the number of Gaussians. We report the best metrics for each baseline here (refer Appendix for ablation). Our opacity regularization results in 
$3.5\times$ to $5\times$ reduction in Gaussian count with nearly identical performance as the larger models. Similar level of compression 
is achieved only by the gradient threshold baseline that reduces densification.
However, it results in a large drop in performance. \\

\begin{table}[b]
\vspace{-0.2in}
    \begin{minipage}{.60\linewidth}
    \caption{\textbf{Breakdown of memory usage in CompGS.} We observe that just $4$ non-quantized values of the total $59$ values per Gaussian contribute to $68\%$ and $81\%$ of the total memory in our $16$-bit and $32$-bit variants respectively. For the quantized parameters, nearly the entire memory is used to store the indices.}
    \label{tab:kmeans_breakdown}
    \begin{minipage}{.55\linewidth}
    % \vspace{-0.2in}
    \centering
        \scalebox{0.8}{
        \begin{tabular}{lcc}
        \toprule
             & Non & Quant \\
             & Quant        \\
            \midrule
            Num Params   & 4  & 55 \\
            Mem (16bit) & 68\% & 32\%\\
            Mem (32bit) & 81\% & 19\% \\
        \bottomrule
        \end{tabular}}
        % \caption{Caption}
        % \label{tab:my_label}
    \end{minipage}
    \hspace{0.01in}
    \begin{minipage}{.40\linewidth}
        % \centering
        % \vspace{-0.1cm}
        \scalebox{0.9}{
        \begin{tabular}{cc}
        \toprule
             \multicolumn{2}{c}{k-Means Quantization}\\
             Index & Codebook \\
            \midrule
            99\% & 1\%\\
            98\% & 2\% \\
        \bottomrule
        \end{tabular}}
    \end{minipage}
    \end{minipage}
    \hspace{0.1in}
    \begin{minipage}{.40\linewidth}
    \centering
    \caption{\textbf{Compression performance tradeoff.} Gaussian count decreases drastically with heavy regularization but also results in some drop in performance on \mipnerf dataset. We choose $\lambda_\textrm{reg}=10^{-7}$ as default.} % setting.}
    %\vspace{-0.1in}
    \scalebox{0.85}{
    \begin{tabular}{ccccc}
    \toprule
         $\lambda_\textrm{reg}(\times 10^{-7})$ & SSIM & PSNR & LPIPS & \#Gauss\\
     \midrule
         $0.5$ & $\bm{0.808}$ & $\bm{27.17}$ & $\bm{0.234}$ & 1.21M \\ 
         $1.0$ & 0.806 & 27.12 & 0.240 & 845K \\
         $2.0$ & 0.801 & 26.98 & 0.253 & 536K \\
         $3.0$ & 0.794 & 26.83 & 0.266 & $\bm{390K}$ \\
    \bottomrule
    \end{tabular}
    }
    \label{tab:perf_vs_mem}
    \end{minipage}
    \vspace{-0.2in}
\end{table}

\noindent \textbf{Results on \arkit and DL3DV datasets: } 
Table~\ref{tab:kmeans_arkit} % and Fig.~\ref{fig:arkit} 
shows the quantitative and qualitative results on our large-scale \arkit benchmark. Our compressed model achieves
nearly the same performance as \gauss with ten times smaller memory. Unlike \ours, the \dc method suffers a 
significant drop in quality. We also report PSNR-AM as the PSNR calculated using arithmetic mean of MSE over all the scenes in the dataset to prevent the domination of high-PSNR scenes. Similarly, Table~\ref{tab:kmeans_dl3dv} shows the performance of \ours with both KMeans quantization and opacity regularization. \ours achieves nearly $30\times$ compression compared to 3DGS with a small drop in performance.

\vspace{-.1in}
\subsection{Ablations}
\vspace{-.1in}
We analyze our design choices and the effect of various hyperparameters on reconstruction performance and model size. \\

\vspace{-.1in}
\noindent \textbf{Memory break-down of CompGS:} In Table~\ref{tab:kmeans_breakdown}, we show the
contribution of various components to the final memory usage of \ours. Out of $59$ parameters of each
Gaussian, we quantize $55$ parameters of color and covariance while the $3$ position and $1$ opacity
parameters are used as is. However, the bulk of the stored memory ($68\%$ and $81\%$ for 16- and 32-bits) 
is due to the non-quantized parameters. For the quantized parameters, nearly all the memory is used to
store the cluster assignment indices with less than $2\%$ used for the codebook. \\

\begin{table}[t]
%\vspace{-0.1in}
\begin{minipage}{0.42\linewidth}
    \caption{\textbf{Performance and training time trade-off.} Depending on user's needs, it is possible to obtain models with fast training or high performance. The hyperparameters of vector quantization - number of K-Means iterations (iters), K-Means index assignment frequency (freq) and codebook size (\#codes) can be varied to obtain the desired point on the curve. They offer a good trade-off, with huge decrease in training time with minor changes in performance. Results are shown on MipNerf-360.}
    \label{tab:perf_vs_time}
    \vspace{-0.1in}
    \centering
    \scalebox{0.9}{
    \begin{tabular}{ccc|ccc}
    \toprule
         Iters & Freq & \#Codes & SSIM & PSNR & Time \\ % & LPIPS
     \midrule
         1  & 100 & 8K  & 0.802 & 26.94 & 19.3\\  % & 0.246 
         3  & 100 & 8K  & 0.802 & 26.94 & 20.9\\  % & 0.245 
         5  & 100 & 8K  & 0.802 & 26.95 & 22.5\\  % & 0.245 
         10 & 100 & 8K  & 0.802 & 26.95 & 26.5\\  % & 0.245 
         \midrule
         5  &  50 & 8K  & 0.803 & 27.00 & 28.7\\  % & 0.244 
         % 5  & 100 & 8K  & 0.802 & 26.95 & 22.5\\  % & 0.245 
         5  & 200 & 8K  & 0.799 & 26.76 & 19.4\\  % & 0.251 
         5  & 500 & 8K  & 0.783 & 26.15 & $\bm{18.1}$\\  % & 0.276 
         \midrule
         5  & 100 & 4K  & 0.800 & 26.84 & 19.6\\  % & 0.248 
         5  & 100 & 16K & 0.804 & 27.05 & 28.9\\  % & 0.243 
         5  & 100 & 32K & $\bm{0.806}$ & $\bm{27.12}$ & 42.4\\  % & 0.240 
    \bottomrule
    \end{tabular}
    }
    % \vspace{-0.2in}
\end{minipage}
\hspace{0.1in}
\begin{minipage}{0.58\linewidth}
% \begin{table}[]
    % \vspace{-0.1in}
    \centering
    \caption{\textbf{Effect of quantization on different Gaussian parameters.} Each Gaussian in \gauss is parameterized using position (pos), scale, rotation (rot) and color (DC and harmonics SH). We analyze the effect of quantizing combinations of these parameters on the view synthesis performance. SH+DC denotes that a single codebook is used for both SH and DC. Position values cannot be quantized without greatly affecting model performance. The rest of the parameters can be simultaneously combined to obtain a high degree of compression without much loss in quality.} % of the generated views.}
    \label{tab:kmeans_variants}
    \vspace{-0.1in}
    \scalebox{0.8}{
    % \resizebox{\textwidth}{!}{
    \begin{tabular}{lcc|cc|c}
    \toprule
         Quantized  & \multicolumn{2}{c}{Train} & \multicolumn{2}{c}{Truck}\\
         Params & SSIM$^\uparrow$ & PSNR$^\uparrow$ & SSIM$^\uparrow$ & PSNR$^\uparrow$ &  Mem \\
        \midrule
        \gauss                  & 0.811 & 21.99 & 0.878 & 25.38 & 20.0 \\ % & 21.2 \\
        \dc                     & 0.798 & 21.40 & 0.871 & 24.92 & 4.8 \\ % & 5.0  \\ 
        \midrule
        \rowcolor{ourcolor}
        \multicolumn{6}{c}{Variants of \ours} \\
        \midrule
        Pos                     & 0.673 & 19.81 & 0.730 & 21.65 & 19.0  \\ % & 19.7 \\
        SH                      & 0.809 & 21.88 & 0.876 & 25.27 & 4.8  \\ % & 5.0  \\
        SH, DC                  & 0.806 & 21.68 & 0.875 & 25.24 & 3.8  \\ % & 4.0 \\
        Rot(R)                  & 0.808 & 21.83 & 0.876 & 25.32 & 18.7  \\ % & 19.7 \\
        Scale(Sc)               & 0.809 & 21.79 & 0.877 & 25.30 & 19.0  \\ % & 20.1 \\
        SH,R                    & 0.805 & 21.67 & 0.874 & 25.20 & 3.5  \\ % & 3.7\\
        SH,Sc                   & 0.806 & 21.63 & 0.875 & 25.18 & 3.8  \\ % & 4.1\\
        SH,Sc,R                 & 0.801 & 21.64 & 0.872 & 25.02 & 2.6  \\ % & 2.8 \\   % -84.3\% & -34.9\% & \\
        SH+DC,Sc,R              & 0.797 & 21.41 & 0.868 & 24.89 & 1.6  \\ % & 1.7 \\% -89.0\% & -53.7\%\\  
        SH,DC,Sc,R              & 0.801 & 21.64 & 0.871 & 24.97 & 1.7  \\ % & 1.8 \\ % -88.7\% & -52.3\%\\
        SH,DC,Sc,R Int16        & 0.790 & 21.49 & 0.869 & 24.93 & 1.0  \\ % & 1.8 \\ % -88.7\% & -52.3\%\\
        \bottomrule
    \end{tabular}
    }
% \end{table}
    \end{minipage}
    \vspace{-.2in}
\end{table}
% \end{figure}

\noindent \textbf{Trade-off between performance, compression, and training time:}
Compressing the Gaussian parameters 
comes with a trade-off, particularly between performance and training time. In our method, the size of codebook, frequency of code assignment and number of iterations in
code computation control this trade-off. Similarly, regularization strength can be modified
in Gaussian count reduction to obtain a trade-off between performance and compression.
We show ablations on these hyperparameters in Tables~\ref{tab:perf_vs_mem} and~\ref{tab:perf_vs_time}. 
\ours offers great flexibility, with different levels of compression and training time without sacrificing
much on performance. \\

\noindent \textbf{Parameter selection for quantization:}
Table~\ref{tab:kmeans_variants} shows the effect of quantizing different subsets of the Gaussian parameters on the 
\tandt dataset. Quantizing the position parameters significantly reduces the performance on both the 
scenes. We thus do not quantize position in any of our other experiments. Quantizing only the 
harmonics (SH) of color parameter is nearly identical in size to the no-harmonics (\dc) of 
\gauss. Our SH has very little drop in metrics compared to \gauss while \dc is much worse off without
the harmonics. As more parameters are quantized, the performance of \ours slowly reduces. The 
combination of all color and covariance parameters still results in a model with good qualitative and
quantitative results.\\

\begin{figure}[t]
\begin{minipage}{0.5\linewidth}
    \vspace{-.4in}
\begin{center}
  \includegraphics[width=0.8\linewidth]{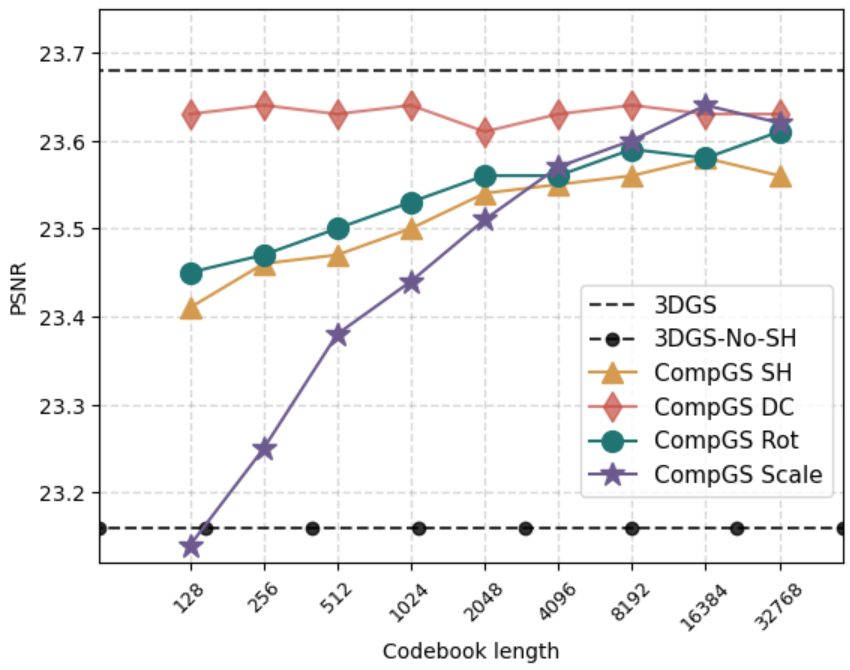}
\end{center}
    \vspace{-.2in}
   \caption{\textbf{Effect of codebook length.} We vary codebook size while quantizing one of the paramteres (SH, DC, Rotation, Scale).} %parameters by quantizing just one of them.} 
   \label{fig:ab_ncls}
  \vspace{-.5in}
% \end{wrapfigure}
    \end{minipage}
    \hspace{0.05in}
    \begin{minipage}{0.5\linewidth}
% \begin{table*}[]
    \centering
    \scalebox{0.85}{
    % \resizebox{\linewidth}{!}{
    \begin{tabular}{lccc}
    \toprule
    Dataset & \multicolumn{3}{c}{Mip-NeRF360} \\
    Method & SSIM$^\uparrow$ & PSNR$^\uparrow$ & LPIPS$^\downarrow$ \\
    \midrule
    \gauss & 0.815 & 27.21 & 0.214  \\
    \gauss$^*$ & 0.813 & 27.42 & 0.217  \\
    \ours 4k & 0.804 & 26.97 & 0.234   \\ 
    \ours Shared Codebook & 0.797 & 26.64 & 0.242 \\
    \bottomrule
    \end{tabular}
    }
    \caption{\textbf{Effect of shared codebook.} A frozen codebook trained on one scene (`Counter' scene) generalizes well to all other scenes in MipNerf-360 dataset. Only code assignments are learnt during training.}
    \label{tab:shared_codebook}
    \end{minipage}
    \vspace{-0.1in}
% \end{table*}
\end{figure}

\vspace{-8pt}
\noindent \textbf{Effect of codebook size:} 
Fig.~\ref{fig:ab_ncls} shows the effect of codebook size for quantization of 
different Gaussian parameters on the \tandt dataset. The DC component of color has 
the smallest drop in performance upon quantization and achieves results similar to the non-quantized version 
with as few as $128$ cluster centers. The harmonics (SH) components of color lead to a much bigger drop at
lower number of clusters and improve as more clusters are added. Note that \ours with only SH components
is nearly the same size as \dc but has better performance ($23.43$ for ours vs. $23.14$ for
\dc). The covariance parameters (rotation and scale) have a drop in performance at a codebook 
size of $1024$ but improve as the codebook size is increased. 

\noindent \textbf{Generalization of codebook across scenes}
We train our train our method  
on a single scene (`Counter') of the \mipnerf dataset. We then freeze the codebook and calculate only assignments for the rest of the eight scenes in the dataset and report the averaged performance metrics over all scenes (Fig.~\ref{tab:shared_codebook}). Interestingly, we observe that the shared codebook generalizes well across all scenes with a small drop in performance compared to learning a codebook for each scene. Sharing learnt codebook can further reduce the memory requirement and can help speed up the training of CompGS. The quality of the codebook can be improved by learning it over multiple scenes. \\

\noindent {\bf Conclusion:}
3D Gaussian Splatting efficiently models 3D radiance fields, outperforming NeRF in learning and rendering efficiency at the cost of increased storage. To reduce storage demands, we apply opacity regularization and K-means-based vector quantization, compressing indices and employing a compact codebook. Our method cuts the storage cost of 3DGS by almost $45\times$, increases rendering FPS by $2.5\times$ while maintaining image quality across benchmarks.\\

\noindent\textbf{Acknowledgments:}
This work is partially funded by NSF grant 1845216 and DARPA Contract No. HR00112190135 and HR00112290115.

{
    \small
    \bibliographystyle{splncs04}
    \bibliography{bibliography}
}

\clearpage

% \onecolumn
\begin{appendices}
\renewcommand{\thesection}{\Alph{section}}
\renewcommand\thefigure{\thesection.\arabic{figure}} 
\renewcommand\thetable{\thesection.\arabic{table}} 
% \appendixpagename
% \begin{document}
\title{Appendix}
\maketitle
% \maketitlesupplementary

Here, we compare the performance of our \ours with state-of-the-art approaches on the NeRF-Synthetic dataset (Section~\ref{sec:synnerf}). 
Section~\ref{sec:shared_codebook} shows exploratory results on generalization of the learnt vector codebook across scenes and 
Section~\ref{sec:code_hist} provides insights on the learnt codebook assignments. We also provide scene-wise results (section~\ref{sec:scene_metrics}), ablations on baselines (section~\ref{sec:ablation_count}) and additional visualizations and qualitative comparisons on the \arkit dataset (section~\ref{sec:add_quals}). 

\section{Results on NeRF-Synthetic dataset}
\label{sec:synnerf}

The results (PSNR) for the NeRF-Synthetic dataset~\cite{mildenhall2020nerf} are presented in Table~\ref{tab:comp_sota_synnerf}. Our CompGS approach achieves an impressive average improvement of $1.13$ points in PSNR compared to the 3DGS-No-SH baseline while using less than half its memory. As reported in the main submission, we report metrics for 3DGS both from the original paper and using our own runs. We observe an improvement for 3DGS~\cite{kerbl20233d} over their official reported numbers by $0.5$ points. 

\begin{table*}[]
    
    \centering
    \caption{\textbf{Results on NeRF-Synthetic dataset.} Here, we present the PSNR values for the synthesized novel views on the NeRF-Synthetic dataset~\cite{mildenhall2020nerf}. Our CompGS approach achieves an impressive average improvement of $1.13$ points in PSNR compared to the 3DGS-No-SH baseline while using less than half its memory. As reported in the main submission, we report metrics for 3DGS both from the original paper and using our own runs. We observe an improvement of 3DGS over the reported numbers by $0.5$points. $^*$ indicates our own run.}
    % \scalebox{0.9}{
    \resizebox{\textwidth}{!}{
    \begin{tabular}{lccccccccc}
    \toprule
                & Mic & Chair & Ship & Materials & Lego & Drums & Ficus & Hotdog & Avg. \\
    \midrule
    Plenoxels   & 33.26 & 33.98 & 29.62 & 29.14 & 34.10 & 25.35 & 31.83 & 36.81 & 31.76 \\
    INGP-Base   & 36.22 & 35.00 & 31.10 & 29.78 & 36.39 & 26.02 & 33.51 & 37.40 & 33.18 \\    
    Mip-Nerf    & 36.51 & 35.14 & 30.41 & 30.71 & 35.70 & 25.48 & 33.29 & 37.48 & 33.09 \\
    Point-NeRF  & 35.95 & 35.40 & 30.97 & 29.61 & 35.04 & 26.06 & 36.13 & 37.30 & 33.30 \\
    \gauss      & 35.36 & 35.83 & 30.80 & 30.00 & 35.78 & 26.15 & 34.87 & 37.72 & 33.32 \\
    3DGS$^*$  & 36.80 & 35.51 & 31.69 & 30.48 & 36.06 & 26.28 & 35.49 & 38.06 & 33.80 \\
    \dc& 34.37 & 34.09 & 29.86 & 28.42 & 34.84 & 25.48 & 32.30 & 36.43 & 31.97 \\
    \rowcolor{ourcolor}
    \ours 4k    & 35.99 & 34.92 & 31.05 & 29.74 & 35.09 & 25.93 & 35.04 & 37.04 & 33.10 \\  
    \bottomrule
    \end{tabular}
    }
    \label{tab:comp_sota_synnerf}
\end{table*}

\section{Generalization of codebook across scenes}
\label{sec:shared_codebook}

We train our vector quantization approach including the codebook and the code assignments on a single scene (`Counter') of the \mipnerf dataset. We then freeze the codebook and learn only assignments for the rest of the eight scenes in the dataset and report the averaged performance metrics over all scenes. In addition to the results in Fig.5 of the main submission, we provide results with our 32K variant here in table~\ref{tab:app_shared_codebook}. Interestingly, we observe that the shared codebook generalizes well across all scenes with a small drop in performance compared to learning a codebook for each scene. Sharing learnt codebook can further reduce the memory requirement and can help speed up the training of CompGS. The quality of the codebook can be improved by learning it over multiple scenes. Fig.~\ref{fig:fixed_centers} shows qualitative comparison of the same. There are no apparent differences between \ours and CompGS-Shared-Codebook approaches. 

\begin{table*}[]
    
    \centering
    \caption{\textbf{Effect of shared codebook.} We train our vector quantization approach including the codebook and the code assignments on a single scene (`Counter') of the \mipnerf dataset. We then freeze the codebook and learn only assignments for the rest of the eight scenes in the dataset and report the averaged performance metrics over all scenes. Interestingly, we observe that the shared codebook generalizes well across all scenes with a small drop in performance compared to learning a codebook for each scene. Sharing learnt codebook can further reduce the memory requirement and can help speed up the training of CompGS. The quality of the codebook can be improved by learning it over multiple scenes.}
    % \scalebox{0.9}{
    % \resizebox{\linewidth}{!}{
    \begin{tabular}{lccc}
    \toprule
    Dataset & \multicolumn{3}{c}{Mip-NeRF360} \\
    Method & SSIM$^\uparrow$ & PSNR$^\uparrow$ & LPIPS$^\downarrow$ \\
    \midrule
    \gauss & 0.815 & 27.21 & 0.214  \\
    \gauss$^*$ & 0.813 & 27.42 & 0.217  \\
    \ours 4K & 0.804 & 26.97 & 0.234   \\ 
    \ours 32K & 0.806 & 27.12 & 0.240  \\
    \ours Shared Codebook 4K & 0.797 & 26.64 & 0.242 \\
    \ours Shared Codebook 32K & 0.800 & 26.780 & 0.247 \\
    \bottomrule
    \end{tabular}
    % }
    \label{tab:app_shared_codebook}
\end{table*}

\section{Analysis of learnt code assignments}
\label{sec:code_hist}

\begin{figure*}
    \centering    
    \includegraphics[width=1\linewidth]{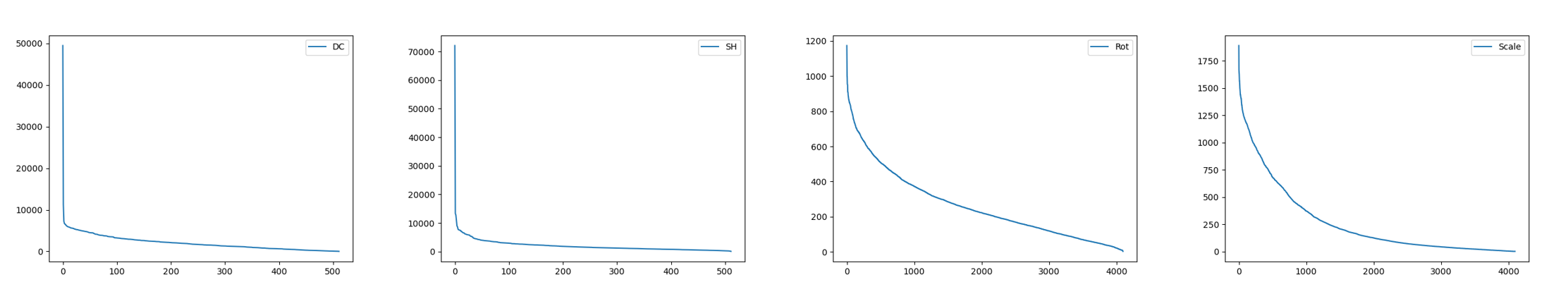}
    \caption{\textbf{Histogram of code assignments.} We plot the sorted histogram of the code assignments (cluster to which each Gaussian belongs to) for each parameter on the `Train' scene of \tandt dataset. We observe that just a single code out of the $512$ in total is assigned to nearly $5\%$ of the Gaussians for both the SH and DC parameters. Similarly, a few clusters dominate even in the case of rotation and scale parameters, albeit to a lower extent. Such a non-uniform distribution of cluster sizes suggest that further compression can be achieved by using Huffman coding to store the assignment indices.}
    \label{fig:hist_cnts_kmeans}
\end{figure*}

In Fig.~\ref{fig:hist_cnts_kmeans}, we plot the sorted histogram of the code assignments (cluster to which each Gaussian belongs to) for each parameter on the `Train' scene of \tandt dataset. We observe that just a single code out of the $512$ in total is assigned to nearly $5\%$ of the Gaussians for both the SH and DC parameters. Similarly, a few clusters dominate even in the case of rotation and scale parameters, albeit to a lower extent. Such a non-uniform distribution of cluster sizes suggest that further compression can be achieved by using Huffman coding to store the assignment indices.

\section{Scene-wise Metrics}
\label{sec:scene_metrics}
For brevity, we reported the averaged metrics over all scenes in a dataset in our main submission. Here in table~\ref{tab:scene_metrics}, 
we provide the detailed scene-wise metrics for MipNerf-360, Tanks and Temples and DeepBlending datasets. 

\begin{table}[]
    \centering
    \caption{\textbf{Scene-wise metrics.} We report the scene-wise metrics on all the scenes for both 3DGS and \ours 32K. As observed in the averaged metrics in the main submission, \ours achieves a high level of compression and fast rendering without losing much on rendering quality.}
    \label{tab:scene_metrics}
    \begin{tabular}{ccccccccc}
    \toprule
        Scene & Method & SSIM$^\uparrow$ & PSNR$^\uparrow$ & LPIPS$^\downarrow$ & Train Time(s) & FPS & Mem(MB) & \#Gauss\\
        \midrule
        \multirow{2}{*}{Bicycle}  & 3DGS & 0.763 & 25.169 & 0.212 & 1845 & 68  & 1422 & 6026079 \\
                                  & \ours 32K & 0.755 & 25.068 & 0.244 & 2123 & 242 & 29   & 1314018  \\
                                  \midrule
        \multirow{2}{*}{Bonsai}   & 3DGS & 0.940 & 31.918 & 0.206 & 876  & 276 & 293    & 1241520    \\
                                  & \ours 32K & 0.932 & 31.195 & 0.223 & 1405 & 492 & 10   & 377227 \\
                                  \midrule
        \multirow{2}{*}{Counter}  & 3DGS & 0.906 & 29.018 & 0.202 & 968  & 208 & 283    & 1200091 \\
                                  & \ours 32K & 0.895 & 28.467 & 0.222 & 1571 & 356 & 10   & 385515  \\
                                  \midrule
        \multirow{2}{*}{Flowers}  & 3DGS & 0.602 & 21.456 & 0.339 & 1192 & 133 & 851    & 3605827\\
                                  & \ours 32K & 0.588 & 21.262 & 0.367 & 1744 & 343 & 23   & 1037132  \\
                                  \midrule
        \multirow{2}{*}{Garden}   & 3DGS & 0.863 & 27.241 & 0.108 & 1870 & 76  & 1347 & 5709543   \\
                                  & \ours 32K & 0.848 & 26.822 & 0.140 & 2177 & 258 & 30   & 1370624  \\
                                  \midrule
        \multirow{2}{*}{Kitchen}  & 3DGS & 0.926 & 31.510 & 0.127 & 1206 & 158 & 425    & 1801403\\
                                  & \ours 32K & 0.918 & 30.774 & 0.142 & 1729 & 317 & 13   & 564382  \\
                                  \midrule
        \multirow{2}{*}{Room}     & 3DGS & 0.917 & 31.346 & 0.221 & 1020 & 190 & 364    & 1541909\\
                                  & \ours 32K & 0.912 & 31.131 & 0.235 & 1439 & 444 & 9     & 327191 \\
                                  \midrule
        \multirow{2}{*}{Stump}    & 3DGS & 0.772 & 26.651 & 0.215 & 1445 & 104 & 1136  & 4815087 \\
                                  & \ours 32K & 0.770 & 26.605 & 0.236 & 1936 & 303 & 27   & 1226044  \\
                                  \midrule
        \multirow{2}{*}{Treehill} & 3DGS & 0.633 & 22.504 & 0.327 & 1213 & 122 & 879    & 3723675\\
                                  & \ours 32K & 0.634 & 22.747 & 0.355 & 1744 & 336 & 23   & 1002290  \\
                                  \midrule
        \multirow{2}{*}{Train} & 3DGS & 0.811 & 21.991 & 0.209 & 563  & 253 & 254 & 1077461\\
                                  & \ours 32K & 0.804 & 21.789 & 0.231 & 1169 & 456 & 12 & 500811 \\
                                  \midrule
        \multirow{2}{*}{Truck} & 3DGS & 0.878 & 25.385 & 0.148 & 897  & 159 & 611 & 2588966\\
                                  & \ours 32K & 0.872 & 25.092 & 0.165 & 1306 & 494 & 13 & 540081 \\
                                  \midrule
        \multirow{2}{*}{DrJohnson} & 3DGS & 0.898 & 29.089 & 0.247 & 1312 & 121 & 772 & 3270679\\
                                  & \ours 32K & 0.906 & 29.445 & 0.249 & 1717 & 379 & 17 & 714902 \\
                                  \midrule
        \multirow{2}{*}{Playroom} & 3DGS & 0.901 & 29.903 & 0.246 & 1003 & 181 & 551 & 2335846\\
                                  & \ours 32K & 0.908 & 30.347 & 0.253 & 1422 & 589 & 10 & 393414 \\

    \bottomrule
    \end{tabular}
\end{table}

\section{Ablations for Gaussian count reduction}
\label{sec:ablation_count}
We perform ablations to choose the right hyperparameters for the baseline approaches to reduce the Gaussian count. The metrics for the chosen settings are reported in table 3 of the main submission. The ablations for minimum opacity, densification interval and end iteration and gradient threshold are shown in tables~\ref{tab:min_op_ablation},~\ref{tab:dense_freq_ablation}, ~\ref{tab:dense_iter_ablation} and ~\ref{tab:grad_thr_ablation} espectively.
Among these baselines, modify gradient threshold provides the best trade-off between model size and performance.

\begin{table*}[t]
    %\vspace{-0.1in}
    \caption{\textbf{Ablation study on opacity threshold}. We changed the default value of $0.005$ for minimum opacity as a baseline approach for compressing 3DGS by reducing the number of Gaussians. Gaussians with opacity values below the threshold are pruned, resulting in smaller models for lower thresholds. Table 3 in main paper shows the results of this experiment when min opacity is $0.1$.}
    \centering
    \scalebox{0.8}{
    \begin{tabular}{l|cccc|cccc|cccc}
    \toprule
               & \multicolumn{4}{c}{\mipnerf} & \multicolumn{4}{c}{\tandt} & \multicolumn{4}{c}{Deep Blending} \\
               & SSIM & PSNR & LPIPS & \#Gauss & SSIM & PSNR & LPIPS & \#Gauss & SSIM & PSNR & LPIPS & \#Gauss  \\ 
        \midrule
        0.05 & 0.810 & 27.308 & 0.230 & 1.93M & 0.839 & 23.508 & 0.194 & 1.04M & 0.902 & 29.542 & 0.251 & 1.47M\\
        0.1  & 0.802 & 27.120 & 0.244 & 1.46M & 0.833 & 23.439 & 0.204 & 780K & 0.902 & 29.504 & 0.255 & 1.01M \\                
        \bottomrule
    \end{tabular}
    }
    % \vspace{-0.3in}
    \label{tab:min_op_ablation}
\end{table*}

\begin{table*}[t]
    %\vspace{-0.1in}
    \caption{\textbf{Ablation study on densification interval}. We modify the densification interval in 3DGS as a baseline approach for compressing 3DGS by reducing the number of Gaussians. Higher intervals results in less frequent densification and thus smaller number of Gaussians. Table 3 in main paper shows the results of this experiment when interval is $500$.}
    \centering
    \scalebox{0.8}{
    \begin{tabular}{l|cccc|cccc|cccc}
    \toprule
               & \multicolumn{4}{c}{\mipnerf} & \multicolumn{4}{c}{\tandt} & \multicolumn{4}{c}{Deep Blending} \\
               & SSIM & PSNR & LPIPS & \#Gauss & SSIM & PSNR & LPIPS & \#Gauss & SSIM & PSNR & LPIPS & \#Gauss  \\ 
        \midrule
        300 & 0.803 & 27.201 & 0.241 & 1.70M & 0.837 & 23.517 & 0.195 & 1.00M & 0.902 & 29.705 & 0.253 & 1.30M \\
        500 & 0.794 & 26.98 & 0.255 & 1.07M & 0.832 & 23.36 & 0.206 & 709K & 0.902 & 29.76 & 0.258 & 844K \\        
        \bottomrule
    \end{tabular}
    }
    \label{tab:dense_freq_ablation}
\end{table*}

\begin{table*}[t]
    %\vspace{-0.1in}
    \caption{\textbf{Ablation study on densification end iteration}. Early or late stopping of densification process impacts the number of Gaussians and the model performance of 3DGS. We report the results with the value set to $3000$ in our main submission (table 3).}
    \centering
    \scalebox{0.8}{
    \begin{tabular}{l|cccc|cccc|cccc}
    \toprule
               & \multicolumn{4}{c}{\mipnerf} & \multicolumn{4}{c}{\tandt} & \multicolumn{4}{c}{Deep Blending} \\
               & SSIM & PSNR & LPIPS & \#Gauss & SSIM & PSNR & LPIPS & \#Gauss & SSIM & PSNR & LPIPS & \#Gauss  \\ 
        \midrule
        5000 & 0.797 & 27.199 & 0.241 & 1.92M & 0.838 & 23.599 & 0.188 & 1.12M & 0.897 & 29.486 & 0.256 & 1.34M \\
        3000 & 0.780 & 27.02 & 0.267 & 1.12M & 0.835 & 23.55 & 0.194 & 810K & 0.896 & 29.42 & 0.264 & 795K \\
        \bottomrule
    \end{tabular}
    }
    % \vspace{-0.3in}
    \label{tab:dense_iter_ablation}
\end{table*}

\begin{table*}[t]
    %\vspace{-0.1in}
    \caption{\textbf{Ablation study on gradient threshold}. We modify the gradient threshold, used as a criterion in the densification process of 3DGS. A higher threshold results in a smaller 3DGS model. Among the baselines considered for reducing Gaussian count, modify gradient threshold provides the best trade-off between model size and performance. Table 3 in main paper shows the results of this experiment for gradient threshold of $0.00045$.}
    \centering
    \scalebox{0.8}{
    \begin{tabular}{l|cccc|cccc|cccc}
    \toprule
               & \multicolumn{4}{c}{\mipnerf} & \multicolumn{4}{c}{\tandt} & \multicolumn{4}{c}{Deep Blending} \\
               & SSIM & PSNR & LPIPS & \#Gauss & SSIM & PSNR & LPIPS & \#Gauss & SSIM & PSNR & LPIPS & \#Gauss  \\ 
        \midrule
        0.00035 & 0.786 & 26.934 & 0.265 & 1.27M & 0.833 & 23.579 & 0.203 & 833K & 0.901 & 29.574 & 0.254 & 1.41M \\
        0.00045 & 0.769 & 26.57 & 0.292 & 809K & 0.825 & 23.31 & 0.217  & 578K & 0.900 & 29.49 & 0.260 & 1.01M \\
        0.00055 & 0.754 & 26.290 & 0.312 & 552K & 0.818 & 23.041 & 0.228 & 433K & 0.899 & 29.544 & 0.265 & 764K \\
        \bottomrule
    \end{tabular}
    }
    % \vspace{-0.3in}
    \label{tab:grad_thr_ablation}
\end{table*}

\section{Qualitative comparison on \arkit dataset.}
\label{sec:add_quals}

Figures~\ref{fig:sample_arkit} and ~\ref{fig:add_viz_arkit} provide qualitative results on the \arkit dataset. 

\begin{figure*}
\begin{center}
  \includegraphics[width=0.9\linewidth]{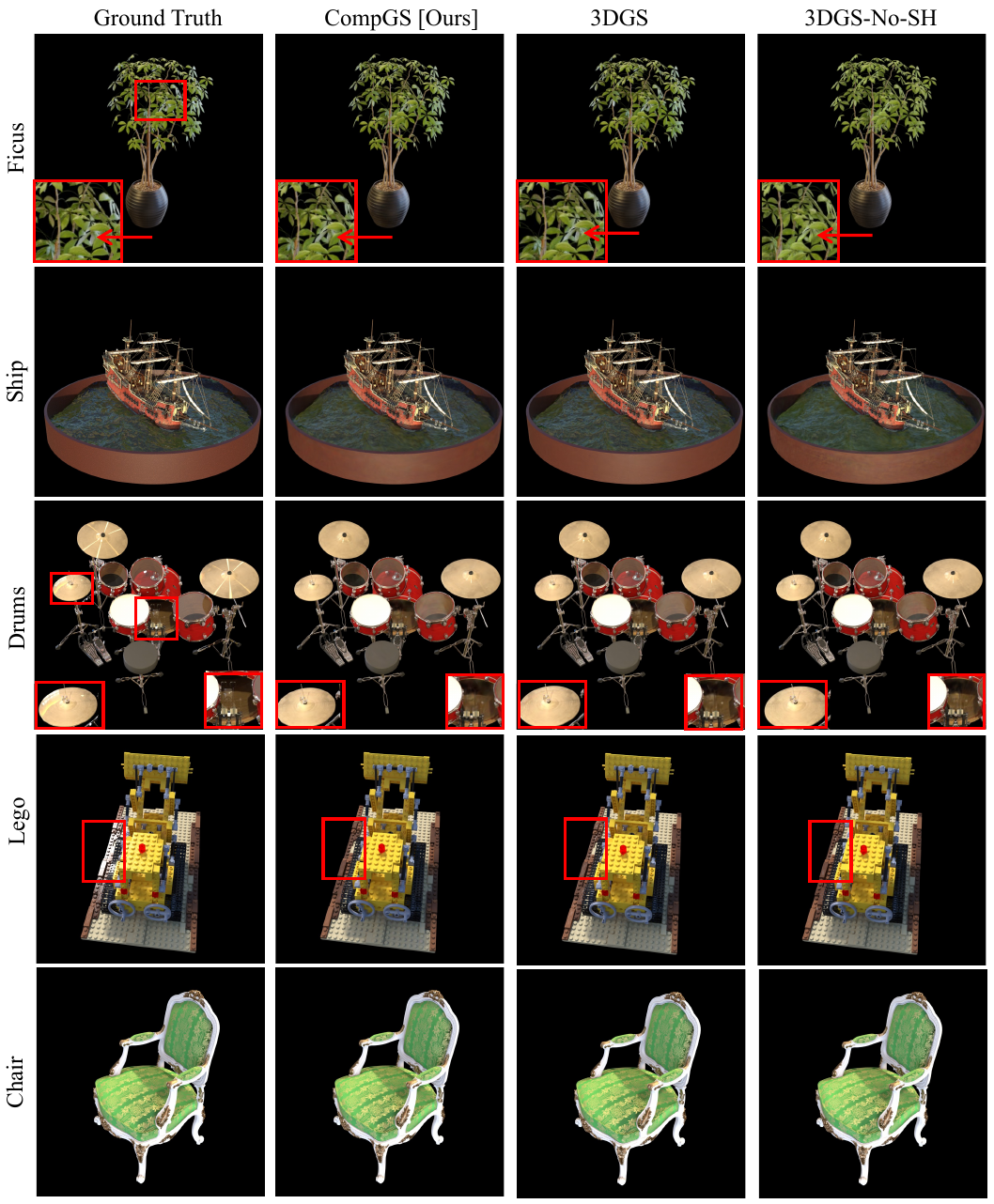}
\end{center}
% \vspace{-.2in}
   \caption{\textbf{Visualization of results on Sythetic-NeRF dataset.} We compare the performance of our compressed CompGS with the original \gauss and \dc approaches on different scenes of the NeRF-Synthetic dataset. The difference between \ours and \dc is apparent in some of these scenes. E.g., \dc fails to effectively model the brown color of branches and shadows and bright light on the leaves of the `Ficus' scene. All approaches including \gauss have imperfect reconstruction in some of the scenes like `Drums' and `Lego'. The scenes and views used for visualization were chosen at random.}
   \label{fig:blender}
 \vspace{-.2in}
\end{figure*}

\begin{figure*}
\begin{center}
  \includegraphics[width=0.9\linewidth]{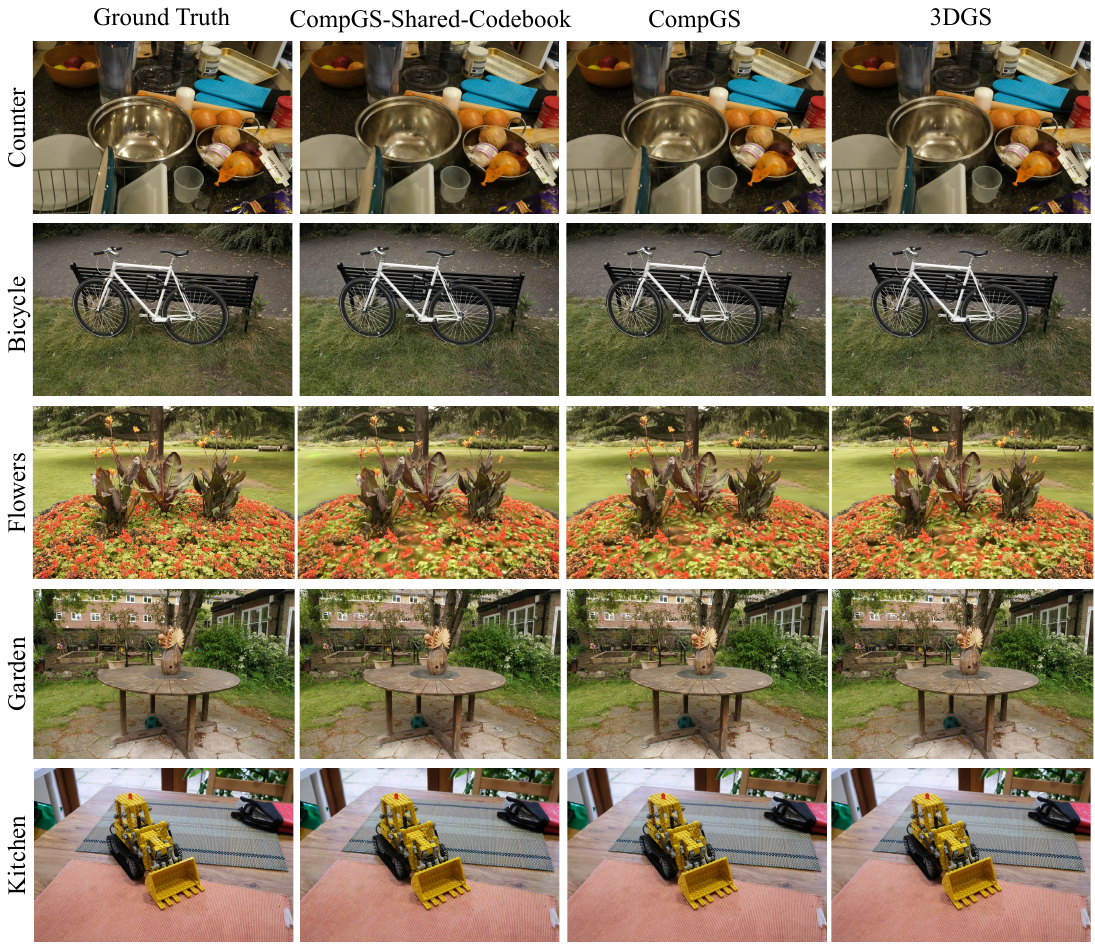}
\end{center}
% \vspace{-.2in}
    \caption{\textbf{Qualitative analysis of shared codebook.} We show the generalization of codebook learned using a single scene on various scenes of the \mipnerf dataset. The codebook was trained on the `Counter' scene (row-1) and frozen for the remaining scenes. The codebooks for all four parameters (DC, SH, Scale, Rot) are shared across scenes. Both \ours and CompGS-Shared-Codebook are visually similar to the uncompressed \gauss with no conspicuous differences between them. \dc requires twice more memory than \ours while \gauss is ten times bigger than \ours. The scenes and views used for visualization were chosen at random.}
   \label{fig:fixed_centers}
 \vspace{-.2in}
\end{figure*}

\begin{figure*}
    \centering    
    \includegraphics[width=1\linewidth]{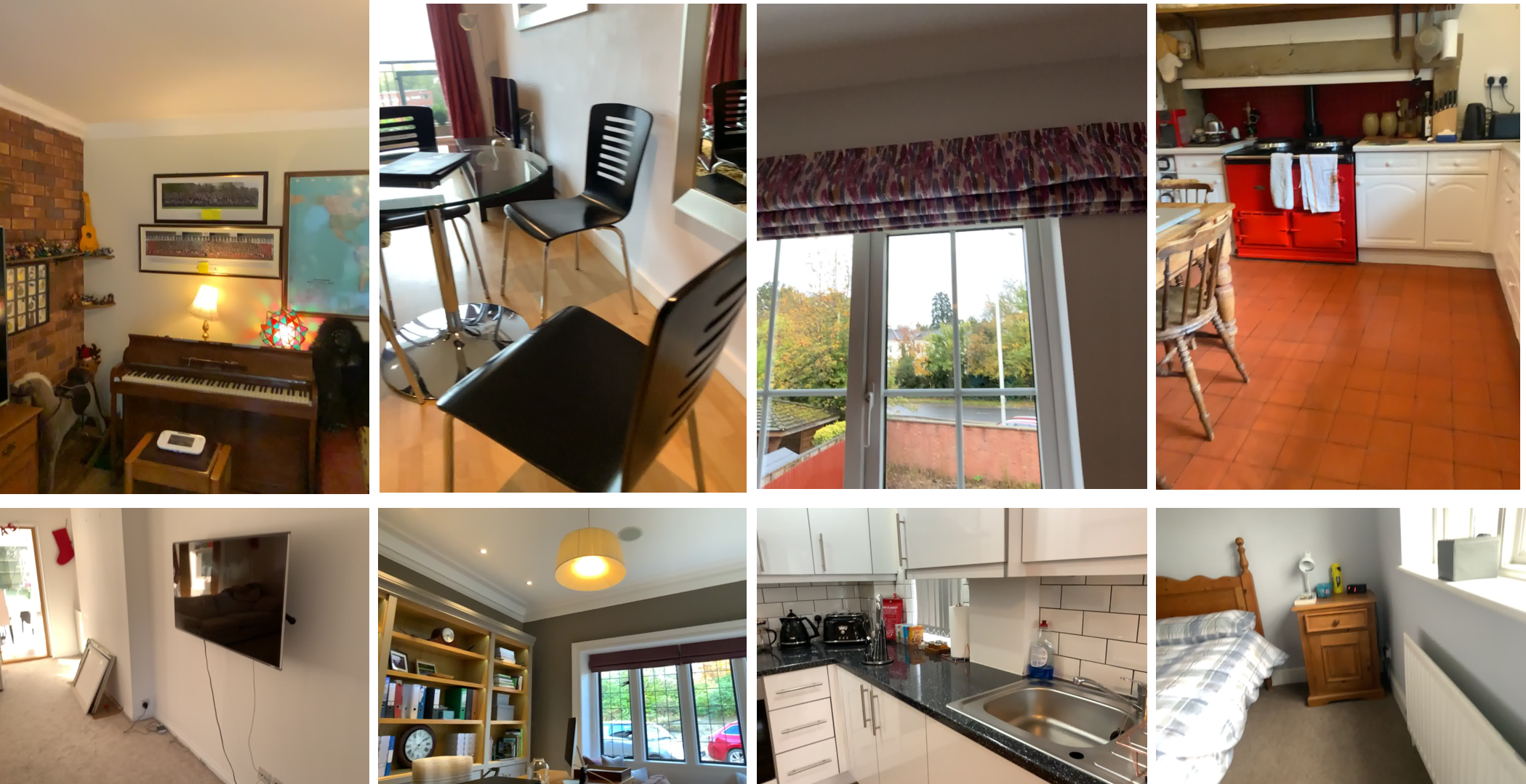}
    \caption{\textbf{Visualization of \arkit dataset.} \arkit is a 3D indoor scene dataset captured using a iPads/iPhones. The dataset consists of videos of indoor environments like houses and office space from multiple view-points. We uniform sample images from each video to form our benchmark dataset for novel view synthesis. Some sample images from different scenes are shown in this figure. The dataset presents unique challenges such as the presence of motion blur due to the use of videos.}
    \label{fig:sample_arkit}
\end{figure*}

\begin{figure*}
    \centering    
    \includegraphics[width=1\linewidth]{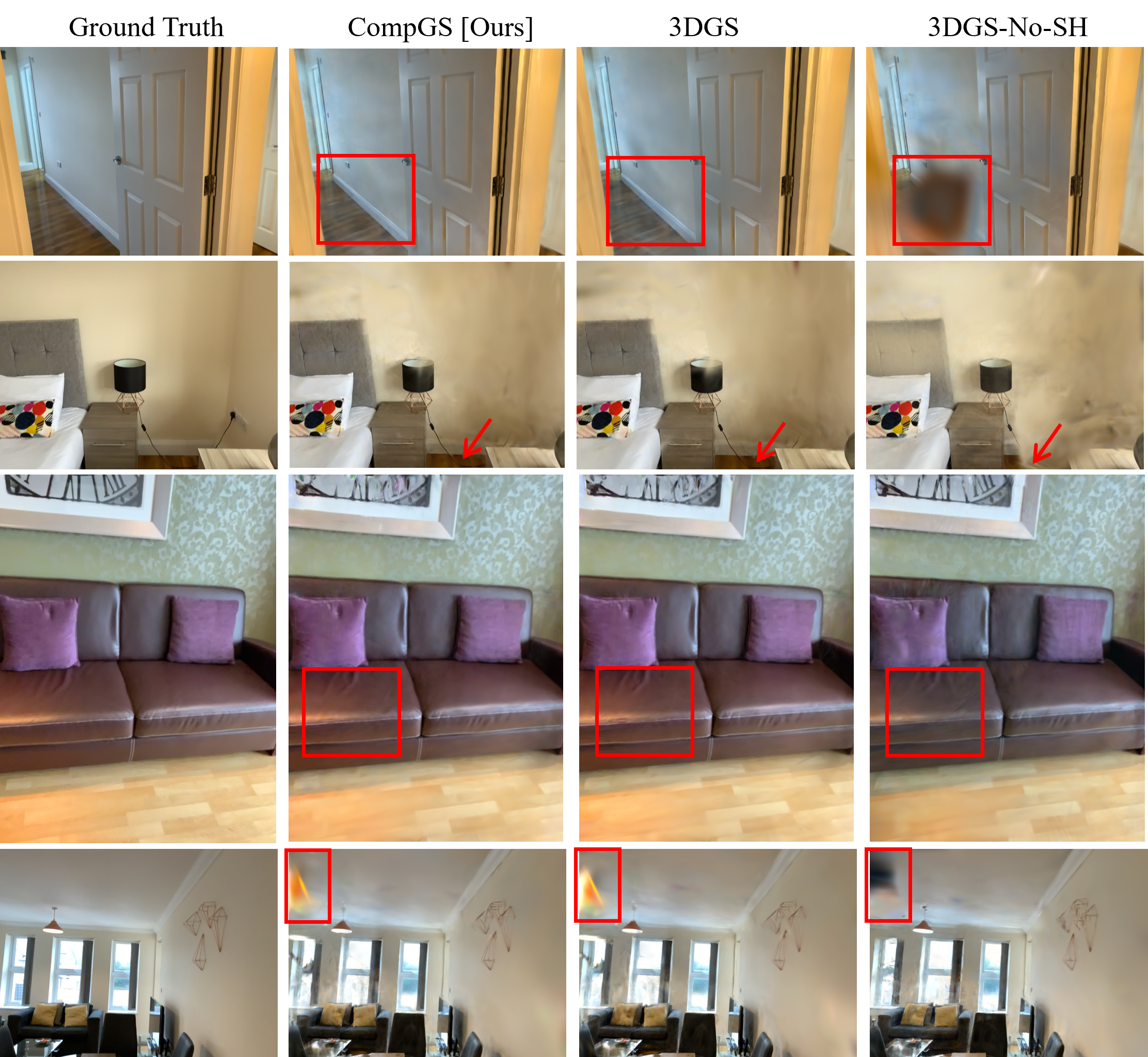}
    \caption{\textbf{Qualitative analysis on \arkit dataset.} We visualize the results of \ours along with the uncomressed \gauss and its variant \dc. Presence of large noisy blobs is a common error mode for \dc on this dataset. It also fails to faithfully reproduce the colors and lighting in several scenes. The visual quality of the synthesized images for all methods is lower on this dataset compared to the scenes present in standard benchmarks like \mipnerf, indicating its utility as a novel benchmark. Further comparison with various NeRF based approaches and more analysis can help improve the results on this dataset.}
    \label{fig:add_viz_arkit}
\end{figure*}

\end{appendices}
\end{document}